\documentclass{article} 
\PassOptionsToPackage{dvipsnames,svgnames,table,xcdraw}{xcolor}
\usepackage{iclr2027_conference,times}

\usepackage[utf8]{inputenc}
\usepackage[T1]{fontenc}
\usepackage{amsmath,amssymb,amsfonts,amsthm}
\usepackage{graphicx,booktabs,multirow,tabularx}
\usepackage{xcolor}
\usepackage{float,sidecap,subcaption}
\usepackage{fontawesome5,nicefrac,microtype,siunitx}
\usepackage{tikz}
\usepackage{pgfplots}
\pgfplotsset{compat=1.18}
\usepgfplotslibrary{groupplots}
\usepackage{placeins}
\usetikzlibrary{positioning,calc}
\usepackage{svg}
\usepackage{algorithm,algpseudocode}
\usepackage{multicol,enumitem,appendix}
\usepackage[most]{tcolorbox}
\usepackage{bm}
\usepackage{hyperref}
\hypersetup{
  colorlinks=true,
  citecolor=magenta,
  pdfborder={0 0 0}
}
\usepackage{url}
\usepackage{xurl} 

\newtheorem*{definition*}{Definition}

\definecolor{pastelblue}{HTML}{E3F2FD}
\definecolor{pastelgreen}{HTML}{E8F5E9}
\definecolor{pastellavender}{HTML}{EDE7F6}
\definecolor{pastelpeach}{HTML}{FFF3E0}
\definecolor{pastelrose}{HTML}{FCE4EC}
\definecolor{pastelmint}{HTML}{E0F2F1}

\newtcolorbox{reviewbox}[2][]{%
  enhanced,
  breakable,
  colback=#2,
  colframe=#2!60!black!20,
  boxrule=0pt,
  arc=4mm,
  outer arc=4mm,
  left=5mm, right=5mm, top=4mm, bottom=4mm,
  shadow={1.5mm}{-1.5mm}{0mm}{black!8},
  attach boxed title to top left={yshift=-3mm, xshift=5mm},
  boxed title style={
    colback=white,
    colframe=white,
    arc=2mm,
    outer arc=2mm,
    boxrule=0pt,
    left=2mm, right=2mm, top=1mm, bottom=1mm,
    shadow={1mm}{-1mm}{0mm}{black!6},
  },
  fonttitle=\normalsize\bfseries,
  coltitle=black,
  title={Human Expert Review~---~#1},
}

\definecolor{artifactbg}{HTML}{ffffff}
\definecolor{artifactframe}{HTML}{d0d0d0}
\definecolor{artifacttitle}{HTML}{333333}

\definecolor{codegreen}{HTML}{228B22}
\definecolor{codeblue}{HTML}{0000CD}
\definecolor{codegray}{HTML}{808080}

\definecolor{promptbg}{HTML}{f9f6f2}
\definecolor{promptframe}{HTML}{d4c4b0}
\definecolor{promptaccent}{HTML}{5a4a3a}

\definecolor{artifactbg}{HTML}{ffffff}
\definecolor{artifactframe}{HTML}{d0d0d0}
\definecolor{artifacttitle}{HTML}{333333}

\definecolor{codegreen}{HTML}{228B22}
\definecolor{codeblue}{HTML}{0000CD}
\definecolor{codegray}{HTML}{808080}

\definecolor{promptbg}{HTML}{f9f6f2}
\definecolor{promptframe}{HTML}{d4c4b0}
\definecolor{promptaccent}{HTML}{5a4a3a}

\usepackage{upquote}
\usepackage{listingsutf8}
\lstdefinestyle{artifactstyle}{
    basicstyle=\ttfamily\scriptsize,
    backgroundcolor=\color{artifactbg},
    commentstyle=\color{codegray}\itshape,
    keywordstyle=\color{codeblue},
    stringstyle=\color{codegreen},
    numberstyle=\tiny\color{codegray},
    breakatwhitespace=false,
    breaklines=true,
    keepspaces=true,
    numbers=left,
    numbersep=8pt,
    showspaces=false,
    showstringspaces=false,
    showtabs=false,
    tabsize=4,
    xleftmargin=12pt,
    framexleftmargin=12pt,
    aboveskip=0pt,
    belowskip=0pt,
    inputencoding=utf8,
    columns=fullflexible,
    upquote=true,
    literate=
        {→}{{$\rightarrow$}}1
        {←}{{$\leftarrow$}}1
        {↔}{{$\leftrightarrow$}}1
        {≤}{{$\leq$}}1
        {≥}{{$\geq$}}1
        {≠}{{$\neq$}}1
        {−}{{-}}1
        {—}{{--}}1
        {–}{{-}}1
        {"}{{\textquotedblleft}}1
        {"}{{\textquotedblright}}1
        {'}{{\textquoteleft}}1
        {'}{{\textquoteright}}1
        {…}{{...}}1
        {×}{{$\times$}}1
        {÷}{{$\div$}}1
        {±}{{$\pm$}}1
        {∞}{{$\infty$}}1
        {α}{{$\alpha$}}1
        {β}{{$\beta$}}1
        {γ}{{$\gamma$}}1
        {δ}{{$\delta$}}1
        {ε}{{$\epsilon$}}1
        {λ}{{$\lambda$}}1
        {π}{{$\pi$}}1
        {σ}{{$\sigma$}}1
        {∑}{{$\sum$}}1
        {∏}{{$\prod$}}1
        {√}{{$\sqrt{}$}}1
        {∈}{{$\in$}}1
        {∉}{{$\notin$}}1
        {⊂}{{$\subset$}}1
        {⊃}{{$\supset$}}1
        {∩}{{$\cap$}}1
        {∪}{{$\cup$}}1
        {∀}{{$\forall$}}1
        {∃}{{$\exists$}}1
        {¬}{{$\neg$}}1
        {∧}{{$\land$}}1
        {∨}{{$\lor$}}1,
}

\lstdefinestyle{promptstyle}{
    basicstyle=\small\scriptsize,
    backgroundcolor=\color{promptbg},
    breakatwhitespace=false,
    breaklines=true,
    breakindent=0pt,
    breakautoindent=false,
    keepspaces=true,
    numbers=none,
    showspaces=false,
    showstringspaces=false,
    showtabs=false,
    tabsize=4,
    aboveskip=0pt,
    belowskip=0pt,
    xleftmargin=0pt,
    framexleftmargin=0pt,
    extendedchars=true,
    inputencoding=utf8,
    upquote=true,
    literate=
        {→}{{$\rightarrow$}}1
        {←}{{$\leftarrow$}}1
        {↔}{{$\leftrightarrow$}}1
        {≤}{{$\leq$}}1
        {≥}{{$\geq$}}1
        {≠}{{$\neq$}}1
        {−}{{-}}1
        {—}{{--}}1
        {–}{{-}}1
        {"}{{\textquotedblleft}}1
        {"}{{\textquotedblright}}1
        {'}{{\textquoteleft}}1
        {'}{{\textquoteright}}1
        {…}{{...}}1
        {×}{{$\times$}}1
        {÷}{{$\div$}}1
        {±}{{$\pm$}}1
        {∞}{{$\infty$}}1
        {α}{{$\alpha$}}1
        {β}{{$\beta$}}1
        {γ}{{$\gamma$}}1
        {δ}{{$\delta$}}1
        {ε}{{$\epsilon$}}1
        {λ}{{$\lambda$}}1
        {π}{{$\pi$}}1
        {σ}{{$\sigma$}}1
        {∑}{{$\sum$}}1
        {∏}{{$\prod$}}1
        {√}{{$\sqrt{}$}}1
        {∈}{{$\in$}}1
        {∉}{{$\notin$}}1
        {⊂}{{$\subset$}}1
        {⊃}{{$\supset$}}1
        {∩}{{$\cap$}}1
        {∪}{{$\cup$}}1
        {∀}{{$\forall$}}1
        {∃}{{$\exists$}}1
        {¬}{{$\neg$}}1
        {∧}{{$\land$}}1
        {∨}{{$\lor$}}1,
}

\newtcblisting{artifact}[2][python]{%
    enhanced,
    breakable,
    colback=artifactbg,
    colframe=artifactframe,
    boxrule=0.5pt,
    arc=3mm,
    outer arc=3mm,
    left=0mm, right=3mm, top=2mm, bottom=2mm,
    shadow={1.5mm}{-1.5mm}{0mm}{black!10},
    attach boxed title to top left={yshift=-3mm, xshift=5mm},
    boxed title style={
        colback=white,
        colframe=artifactframe,
        arc=2mm,
        outer arc=2mm,
        boxrule=0.5pt,
        left=2mm, right=2mm, top=1mm, bottom=1mm,
    },
    fonttitle=\normalsize\bfseries,
    coltitle=artifacttitle,
    title={#2},
    listing only,
    listing options={
        style=artifactstyle,
        language=#1,
    },
}

\newtcblisting{prompt}[1]{%
    enhanced,
    breakable,
    colback=promptbg,
    colframe=promptframe,
    boxrule=0.5pt,
    arc=3mm,
    outer arc=3mm,
    left=5mm, right=5mm, top=4mm, bottom=4mm,
    shadow={1.5mm}{-1.5mm}{0mm}{black!8},
    attach boxed title to top left={yshift=-3mm, xshift=5mm},
    boxed title style={
        colback=white,
        colframe=promptframe,
        arc=2mm,
        outer arc=2mm,
        boxrule=0.5pt,
        left=2mm, right=2mm, top=1mm, bottom=1mm,
    },
    fonttitle=\normalsize,
    coltitle=promptaccent,
    title={#1},
    listing only,
    listing options={
        style=promptstyle,
    },
}
\newcommand{\mainLSTRL}{\ensuremath{19.03}}
\newcommand{\mainLSTRLShort}{\ensuremath{19.0}}
\newcommand{\mainLSTFKL}{\ensuremath{-3.72}}
\newcommand{\mainLSTFKLShort}{\ensuremath{-3.7}}
\newcommand{\mainLSTRKL}{\ensuremath{-10.92}}
\newcommand{\mainLSTRKLShort}{\ensuremath{-10.9}}
\newcommand{\mainLSTPG}{\ensuremath{1.45}}
\newcommand{\mainLSTPGShort}{\ensuremath{1.4}}
\newcommand{\mainLSTFixed}{\ensuremath{-10.81}}
\newcommand{\mainLSTFixedShort}{\ensuremath{-10.8}}
\newcommand{\mainLSTCRISP}{\ensuremath{10.60}}

\newcommand{\sharedLSTReferenceStep}{885}
\newcommand{\sharedLSTReferenceLength}{853.716}

\graphicspath{{figures/}{./}}
\definecolor{lsdblue}{RGB}{65,105,225}

\definecolor{methodRL}{HTML}{FFB000}
\definecolor{methodCRISP}{HTML}{FFD8A8}
\definecolor{methodSGFKL}{HTML}{6F2C91}
\definecolor{methodSGRKL}{HTML}{A56CC1}
\definecolor{methodPGRKL}{HTML}{D8B4DC}
\definecolor{methodLP}{HTML}{FFD166}
\definecolor{methodFixed}{HTML}{000000}
\newcommand{\methodname}[2]{\textcolor{method#1}{#2}}

\title{Mitigating Length-Scaling Tax with Online Distillation}
\author{
Xu Wan$^{1,*}$ \quad Wenyue Xu$^{2,*}$ \quad Shengjie Zhao$^{2}$ \quad Mingyang Sun$^{3,\dagger}$\\
\normalfont $^1$ByteDance Seed \quad $^2$Tongji University \quad $^3$Peking University\\
\normalfont\small $^*$Equal contribution. \quad $^\dagger$Co-corresponding author.
}
\hypersetup{pdfauthor={Xu Wan, Wenyue Xu, Shengjie Zhao, Mingyang Sun}}

\iclrfinalcopy

\begin{document}
\maketitle
\lhead{Preprint}

\begin{abstract}
Length scaling during reinforcement-learning (RL) post-training is often viewed as a sign of improved reasoning ability, especially on difficult problems, but may also make responses to already-solved problems unnecessarily verbose. We quantify this side effect as the \emph{length-scaling tax} (LST): excess response length on already-solved queries without a commensurate accuracy gain. To mitigate LST, we propose \emph{Length Self-Distillation} (LSD), which routes solved prompts to on-policy distillation and retains the original RL objective for unsolved prompts.
LSD uses an exponential moving average of the online policy as its teacher, requiring no external model. We find that LSD achieves comparable or better performance than RL across multiple variants, while substantially curbing response-length growth on easy queries. LSD reduces LST from \mainLSTRLShort\% to \mainLSTFKLShort\% on single-turn reasoning and from 31.4\% to 13.7\% on multi-turn agentic tasks, demonstrating that LSD effectively preserves concise response patterns on easy queries while supporting efficient exploration on difficult queries during RL post-training.

\end{abstract}

\section{Introduction}
\label{sec:introduction}

Scaling the rollout budget is a common way to improve the performance of large
language models (LLMs) on difficult reasoning tasks. At inference time,
prompting models to think longer, sampling multiple responses, and applying
verifier-guided search can translate additional computation into higher
accuracy
\citep{wei2022chain,wang2022self,lightman2024lets,snell2024scaling}.
Yet the value of this computation depends on problem difficulty. Extended
deliberation and self-correction can help on difficult problems, but offer
little benefit once a problem is already solved. Therefore, adaptively allocating budgets has become an important design axis for modern reasoning models
\citep{muennighoff2025s1,aggarwal2025l1,yang2025qwen3,
openai2025o3,anthropic2025thinking,google2025gemini}.

In parallel, reinforcement learning with verifiable rewards (RLVR) has become
a central post-training mechanism for eliciting LLMs' reasoning and agentic capabilities \citep{shao2024deepseekmath,wanbuffer,jin2025search}.
Since DeepSeek-R1\citep{guo2025deepseek}, the spontaneous growth of response length during RL has often been viewed as a behavioral signature of improving reasoning ability \citep{yeo2025demystifying}. However, a standard RLVR objective jointly optimizes prompts of varying difficulty. Updates that promote longer and more elaborate reasoning on hard problems can also alter the policy’s continuation distribution on easy ones. Under group-relative objectives, this spillover is difficult to correct. Once every response in an easy rollout group is correct, its relative advantages collapse toward zero. The easy group therefore provides no gradient for preserving a concise
solution, while difficult groups continue to reshape the shared policy. 

Moreover, this failure mode may be reinforced by common data-selection strategies.
Dynamic sampling treats all-correct
groups as uninformative and discards them from training
\citep{yu2025dapo}, while difficulty-aware curricula downweight easy problems
and concentrate the training distribution near the policy's competence
frontier
\citep{bae2026online,wanbuffer,qu2026gps}.



Despite the rapidly growing literature on reasoning efficiency, most existing work still characterizes efficiency using coarse-grained aggregate statistics, most notably average response length. A straightforward strategy is to apply stronger length control to easier problems \citep{shen2025dast,xu2026adapthink}. However, such methods do not explicitly preserve the concise behavior that the policy already exhibits on solved queries. Existing evaluations lack a systematic metric for quantifying the unintended lengthening imposed on easy queries by subsequent RL updates. Another line of work focuses on the super-long CoT, especially for difficult problems, because these responses exhibit the most pronounced overthinking behaviors \citep{yuan2026shorten,yi2026shorterbetter,xiang2025just,chen2024not,luo2026o1}. However, imposing length control in this regime often incurs an accuracy cost. Although much of a long trajectory may appear redundant, exploratory branches within that trajectory can still uncover the reasoning path that ultimately leads to the correct solution. Response-level compression may remove not only redundant computation but also reasoning steps necessary to solve the problem.

Complementary to reward shaping, on-policy distillation (OPD) provides dense token-level supervision on states visited by the student \citep{agarwal2024onpolicy}. Because this supervision is evaluated on student-generated prefixes, it directly regularizes the evolving policy along its own state distribution. Recent work has adapted self- and contrastive OPD to reasoning compression, showing that distribution-level supervision can
substantially shorten reasoning while retaining accuracy
\citep{sang2026opsdc,ruan2026contrastive}. However, OPD is
primarily used as a general compression objective. This does not address the asymmetric learning problem considered here. We believe that easy queries need a token-level preservation signal precisely because their relative RL advantage vanishes, whereas difficult queries that remain unsolved should continue to be governed by RLVR.

Motivated by this gap, our objective is to preserve
concise behavior where the policy is already successful without restricting exploration elsewhere. We formalize this training-induced inefficiency as the
\emph{length-scaling tax} (LST): excess response length that emerges on already-solved queries as a shared policy undergoes RL post-training, without a commensurate gain in accuracy. 
Conditioning on a fixed easy query set distinguishes LST from global measures of overthinking while avoiding the survivorship bias that arises from repeatedly redefining the easy set.

To better understand LST, we first conduct several empirical studies to characterize
its behavior and identify its drivers. We find that LST persists across
multiple easy sets defined at different checkpoints, rather than arising from a particular model snapshot. Moreover, we find that training distributions concentrated on hard prompts further amplify the tax.

Based on this, we propose \emph{length self-distillation} (LSD).
At each training step, LSD uses the current rollout accuracy to route solved
prompt groups to an OPD objective, while retaining the original RLVR objective
for unsolved groups. In practice,
LSD requires neither a stronger external teacher nor a separately prompted
concise model. Instead, its teacher is a delayed version of the same policy
lineage, instantiated as a rolling
exponential moving average checkpoint. Within LSD, we compare supervised-gradient forward- and reverse-KL objectives with a
sampled-action policy-gradient estimator of reverse KL, and analyze how these objectives constrain the policy at different levels of granularity.

Our study makes three contributions. 

First, we introduce LST, a query-conditional metric that measures
excess response length on already-solved prompts relative to an
accuracy-qualified reference. We also establish the prevalence of LST, characterize its behavioral signatures, and identify its key drivers.

Second, we propose LSD, a mixed RL and
distillation algorithm that routes solved rollout groups to self-distillation
while retaining the original RLVR objective for unsolved groups. We develop
three complementary implementations and analyze their
different levels of policy-control granularity.

Third, we evaluate LSD on single-turn reasoning and multi-turn agentic tasks, showing that LSD with SG-FKL matches or improves average Pass@1 over RL while reducing LST from \mainLSTRLShort\% to \mainLSTFKLShort\% on single-turn reasoning and from 31.4\% to 13.7\% on multi-turn agentic tasks. We also analyze how the distillation objective, teacher half-life, and routing threshold affect the trade-off between preserving efficient behavior and acquiring new capabilities.

\section{The Length-Scaling Tax}
\label{sec:lst}

Let $\mathcal{X}_{\mathrm{eval}}$ denote the evaluation query set, and let
$x\in\mathcal{X}_{\mathrm{eval}}$ be a query. At each checkpoint $k$ during RL post-training, we sample $N$ responses
\begin{equation}
  y_{k,i}(x)\sim\pi_k(\cdot\mid x),
  \qquad i=1,\ldots,N,
\end{equation}
using a fixed decoding configuration and response budget. Let
$r(x,y)\in\{0,1\}$ denote the correctness reward and $\ell(y)$ the number of
response tokens. The empirical solve rate and mean response length of policy
$\pi_k$ on query $x$ are
\begin{equation}
  \widehat{R}(x;\pi_k)
  =
  \frac{1}{N}\sum_{i=1}^{N}r\!\left(x,y_{k,i}(x)\right),
  \qquad
  \widehat{L}(x;\pi_k)
  =
  \frac{1}{N}\sum_{i=1}^{N}\ell\!\left(y_{k,i}(x)\right).
  \label{eq:query-statistics}
\end{equation}

At an RL anchor checkpoint $b$, we define the easy-query set induced by the
anchor policy $\pi_b$ using an evaluation threshold $\tau$, set to $1$
unless otherwise specified:
\begin{equation}
  \mathcal{E}_b
  =
  \left\{
    x\in\mathcal{X}_{\mathrm{eval}}:
    \widehat{R}(x;\pi_b)\ge\tau
  \right\}.
  \label{eq:easy-set}
\end{equation}
Thus, whether a query belongs to $\mathcal{E}_b$ is determined exclusively by
rollouts from $\pi_b$. At the default threshold $\tau=1$, every sampled
rollout for each selected query is correct at the anchor checkpoint.

To avoid survivorship bias, $\mathcal{E}_b$ is frozen after its construction. At every later
checkpoint, we evaluate the same queries in this fixed easy set. For the frozen set $\mathcal{E}_b$, its mean accuracy and response length
under an evaluated policy $\pi_k$ are
\begin{equation}
  R_b(k)
  =
  \frac{1}{|\mathcal{E}_b|}
  \sum_{x\in\mathcal{E}_b}
  \widehat{R}(x;\pi_k),
  \qquad
  L_b(k)
  =
  \frac{1}{|\mathcal{E}_b|}
  \sum_{x\in\mathcal{E}_b}
  \widehat{L}(x;\pi_k).
  \label{eq:acc-length}
\end{equation}
Here, the subscript $b$ specifies which anchor policy defines the query
set, whereas $k$ specifies which policy is being evaluated on that set.

Let $\mathcal{K}_b$ denote all RL checkpoints evaluated on the fixed easy set
$\mathcal{E}_b$ over the full training trajectory, including checkpoints before
anchor $b$. We define an accuracy-constrained reference that captures the
smallest mean token cost observed while maintaining the easy-query accuracy criterion:
\begin{equation}
  L_b^\star
  =
  \min_{\substack{j\in\mathcal{K}_b:\\
  R_b(j)\ge \tau}}
  L_b(j).
  \label{eq:matched-reference}
\end{equation}
For fair comparisons, all methods use the same RL-frozen set $\mathcal{E}_b$
and shared reference $L_b^\star$: the shortest mean response length among
RL checkpoints whose accuracy on this set is at least $\tau$.
We define the normalized length-scaling tax as
\begin{equation}
  \operatorname{LST}_b(k)
  =
  \frac{L_b(k)-L_b^\star}{L_b^\star}.
  \label{eq:lst}
\end{equation}
A positive $\operatorname{LST}_b(k)$ measures the percentage of excess tokens
used by $\pi_k$ relative to this accuracy-qualified reference.
With $\tau=1$, the reference has perfect empirical accuracy, so additional
length cannot correspond to higher observed accuracy than the reference.
When $R_b(k)=1$ as well, LST compares lengths at identical empirical accuracy.
For explicitly reported settings with $\tau<1$, LST instead measures excess
length under an accuracy threshold; it does not by itself establish waste
at identical accuracy.


\paragraph{LST exists under standard RLVR.} We study a standard group-relative RLVR baseline initialized from Qwen3-4B-Base~\citep{yang2025qwen3}. We post-trained the model on the deduplicated DAPO-Math-17K dataset~\cite{yu2026dapo} with a maximum response budget of $4096$ tokens and evaluate checkpoints on AMC 2023, AIME 2025, and AIME 2026. For every query
and checkpoint, we sample $32$ responses using the same decoding
configuration and response budget. 

\begin{figure}[ht]
    \centering
    \includegraphics[width=1\linewidth]{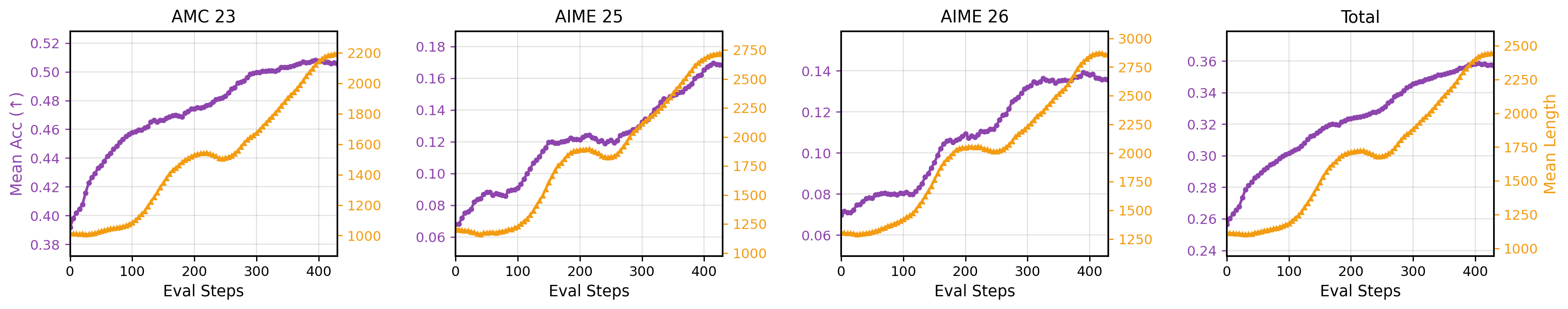}
    \caption{\textbf{Global validation accuracy and response length during RLVR.} Purple circles show mean accuracy, and orange squares show mean response length on three benchmarks and their aggregate over training.}
    \label{fig:rl4k-global-tracking}
\end{figure}

We first examine the dynamics of entire evaluation set in Figure~\ref{fig:rl4k-global-tracking}. As expected, RLVR improves aggregate
accuracy, while mean response length grows steadily throughout training.

\begin{figure}[ht]
  \centering
  \includegraphics[width=\textwidth]
  {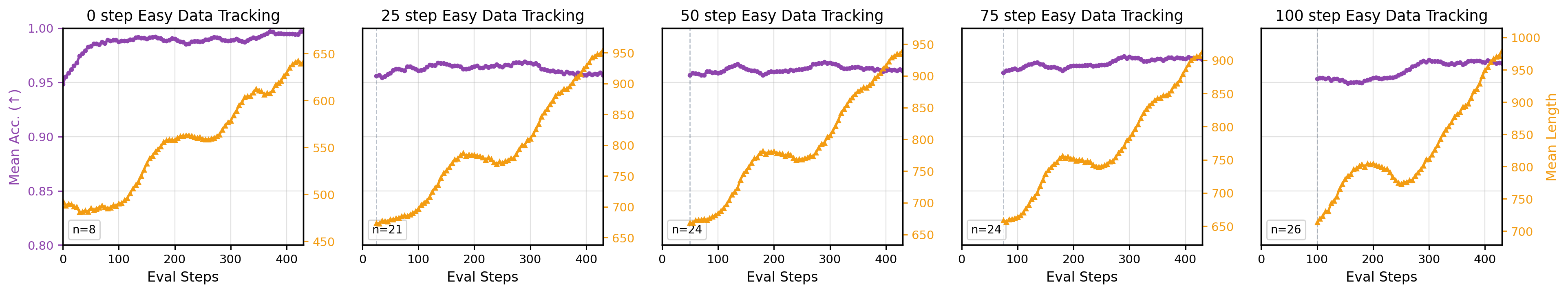}
  \caption{\textbf{Easy-query accuracy saturates while response length keeps
  growing during RLVR.} At each anchor
  $b\in\{0,25,50,75,100\}$, we select queries with solve rate at
  least $\tau=0.875$ and freeze the resulting easy set. }
  \label{fig:rl4k-easy-tracking}
\end{figure}

We next examine whether the length-scaling tax emerges during RLVR by applying the fixed-easy-set protocol in Figure~\ref{fig:rl4k-easy-tracking}. Specifically, at each anchor checkpoint $b\in\{0,25,50,75,100\}$, we select queries with an empirical solve rate of at least $\tau=0.875$, freeze the resulting easy set, and track its accuracy and response length at all subsequent checkpoints. Regardless of which anchor defines the easy set, accuracy remains largely stable, whereas response length continues to increase throughout RLVR.

\begin{figure*}[ht]
  \centering
  \begin{minipage}[ht]{0.48\textwidth}
    \vspace{0pt}
    \paragraph{The extra tokens are not valuable.}
    We further analyze $5{,}824$ responses from the 26 queries in the step-100 easy set $\mathcal{E}_{100}$. Following \citep{xu2026adapthink}, we construct a lexicon of reflection words that capture explicit hesitation, verification, and self-correction. We additionally compute the repeated bigram rate to quantify local phrase repetition within each response. Figure~\ref{fig:easy-behavior} shows that, as training proceeds, the growth in response length is accompanied by a substantial increase in reflection words and repeated bigrams, which is not a desirable behavior for easy queries.
  \end{minipage}
  \hfill
  \begin{minipage}[ht]{0.48\textwidth}
    \vspace{0pt}
    \centering
    \includegraphics[width=\linewidth]{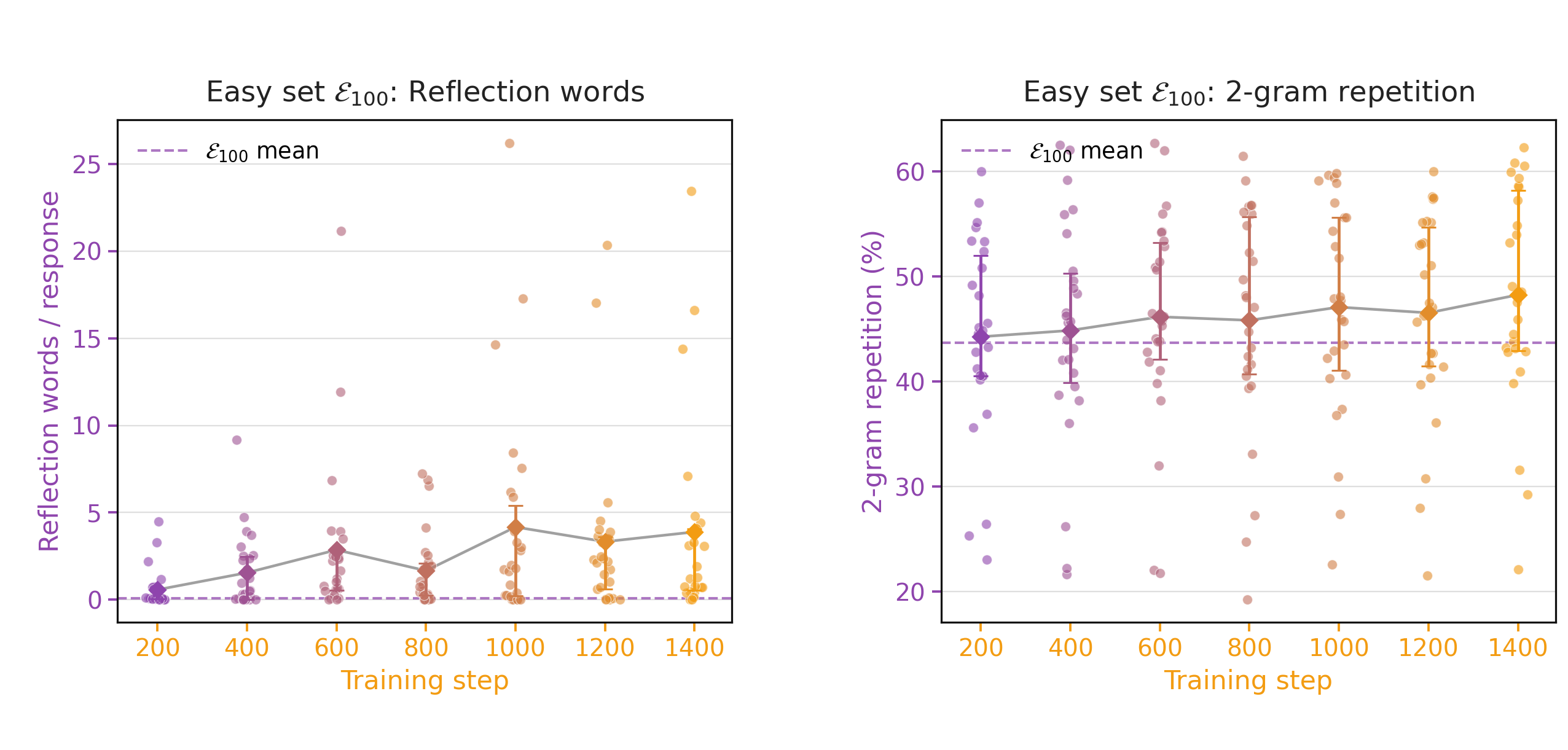}
    \caption{\textbf{Observable behavior on the fixed easyset.} Reflection word frequency and repeated-bigram rate in responses generated by subsequent checkpoints for the same queries in $\mathcal{E}_{100}$.}
    \label{fig:easy-behavior}
  \end{minipage}
\end{figure*}




\section{What Amplifies the Tax?}
\label{sec:diagnosis}

Intuitively, the rollout budget, curriculum design, and training-data difficulty can alter which trajectories contribute gradient signals and how those signals shape the shared policy, thereby affecting the severity of LST. We therefore conduct a series of controlled experiments to test these hypotheses. We refer to the configuration used in Section~2 as \textsc{All-4k}, which trains on the full DAPO-Math-17K mixture with a 4k rollout budget. \textsc{All-8k} uses the same training mixture but doubles the rollout budget to 8k. \textsc{All-4k-8k} first trains with a 4k budget and then continues training the resulting RL checkpoint with an 8k budget. Finally, \textsc{Hard-4k} retains the 4k budget but restricts training to hard prompts that the initial base model solves in at most four out of eight rollouts.

Table~\ref{tab:drivers} reports five anchor-based LST scores and the overall Pass@1 improvement under four training settings.

\noindent\textcircled{\small 1}\enspace\textbf{\textsc{Hard-4k} produces the highest LST across all easy sets.} It suggests that training on harder data causes stronger behavioral spillover to already-solved prompts.

\noindent\textcircled{\small 2}\enspace \textbf{A larger rollout budget amplifies LST when used from the start, but mitigates it when introduced later in training.} At step 240, \textsc{All-8k} has a much higher LST than \textsc{All-4k} across all anchors. A larger budget therefore accelerates LST early in training. The later trend is different. At step 400, \textsc{All-4k-8k} has a lower LST than continued 4k training and achieves a larger Pass@1 improvement. This does not mean that the 8k setting produces shorter responses overall. Its average responses and hard-query responses remain longer, but length growth on easy queries becomes slower.

\begin{table*}[ht]
  \centering
  \caption{\textbf{Controlled comparison of budget and training-data effects.}
  LST is computed with the \textsc{All-4k} fixed-easy reference. Gray rows are
  the corresponding \textsc{All-4k} anchor baselines. Each non-anchor LST entry
  reports the value at $t^\star$, with the colored arrow showing its change from
  the corresponding anchor baseline. Red denotes LST increase and green denotes
  decrease. $\Delta$Pass@1 is full-test-set Pass@1 improvement over the base model.}
  \label{tab:drivers}
  \small
  \resizebox{\textwidth}{!}{%
  \begin{tabular}{lllllllll}
    \toprule
    Setting & Budget & $t^\star$ &
    $\operatorname{LST}_{0}(t^\star)$ &
    $\operatorname{LST}_{25}(t^\star)$ &
    $\operatorname{LST}_{50}(t^\star)$ &
    $\operatorname{LST}_{75}(t^\star)$ &
    $\operatorname{LST}_{100}(t^\star)$ &
    $\Delta$Pass@1 \\
    \midrule
    \rowcolor{gray!10}
    \textsc{All-4k} & 4096 & 240
      & $21.0$
      & $18.1$
      & $16.7$
      & $13.7$
      & $16.3$
      & $+9.1$ pp \\
    \textsc{All-8k} & 8192 & $240^{\ast}$
      & $25.5\,{\scriptscriptstyle\textcolor{red}{\uparrow 4.5}}$
      & $40.9\,{\scriptscriptstyle\textcolor{red}{\uparrow 22.8}}$
      & $38.0\,{\scriptscriptstyle\textcolor{red}{\uparrow 21.3}}$
      & $36.5\,{\scriptscriptstyle\textcolor{red}{\uparrow 22.8}}$
      & $37.4\,{\scriptscriptstyle\textcolor{red}{\uparrow 21.1}}$
      & $+9.7$ pp \\
    \midrule
    \rowcolor{gray!10}
    \textsc{All-4k} & 4096 & 400
      & $42.9$
      & $44.6$
      & $43.0$
      & $40.0$
      & $43.4$
      & $+11.8$ pp \\
    \textsc{All-4k-8k} & $4096\!\rightarrow\!8192$ & $100^{\dagger}$
      & $33.7\,{\scriptscriptstyle\textcolor{green!50!black}{\downarrow 9.2}}$
      & $42.2\,{\scriptscriptstyle\textcolor{green!50!black}{\downarrow 2.4}}$
      & $38.8\,{\scriptscriptstyle\textcolor{green!50!black}{\downarrow 4.2}}$
      & $37.2\,{\scriptscriptstyle\textcolor{green!50!black}{\downarrow 2.8}}$
      & $39.2\,{\scriptscriptstyle\textcolor{green!50!black}{\downarrow 4.2}}$
      & ${+12.6}$ pp \\
    \textsc{All-4k-8k} & $4096\!\rightarrow\!8192$ & $400$
      & $31.4\,{\scriptscriptstyle\textcolor{green!50!black}{\downarrow 11.5}}$
      & $42.6\,{\scriptscriptstyle\textcolor{green!50!black}{\downarrow 2.0}}$
      & $39.3\,{\scriptscriptstyle\textcolor{green!50!black}{\downarrow 3.7}}$
      & $34.8\,{\scriptscriptstyle\textcolor{green!50!black}{\downarrow 5.2}}$
      & $39.3\,{\scriptscriptstyle\textcolor{green!50!black}{\downarrow 4.1}}$
      & $\mathbf{+15.0}$ pp \\
    \textsc{Hard-4k} & 4096 & 400
      & $\mathbf{53.9}\,{\scriptscriptstyle\textcolor{red}{\uparrow 11.0}}$
      & $\mathbf{58.0}\,{\scriptscriptstyle\textcolor{red}{\uparrow 13.4}}$
      & $\mathbf{54.7}\,{\scriptscriptstyle\textcolor{red}{\uparrow 11.7}}$
      & $\mathbf{49.7}\,{\scriptscriptstyle\textcolor{red}{\uparrow 9.7}}$
      & $\mathbf{54.7}\,{\scriptscriptstyle\textcolor{red}{\uparrow 11.3}}$
      & $+11.8$ pp \\
    \bottomrule
  \end{tabular}
  }

  \vspace{0.3em}
  \begin{minipage}{0.98\textwidth}
    \footnotesize
    $^{\ast}$For \textsc{All-8k}, we only report step 240 because the run begins to
    collapse around step 250.
    $^{\dagger}$\textsc{All-4k-8k} continues for 100 steps from the
\textsc{All-4k} step-300 checkpoint, making it comparable to
\textsc{All-4k} at step 400.
  \end{minipage}
\end{table*}

\paragraph{Why outcome-only RL does not correct the drift.}
We explain LST through the policy-gradient signal produced by RLVR. Consider a
rollout group $\{y_i\}_{i=1}^{G}$. Its group-relative advantages vanish when
all responses receive the same correct reward:
\begin{equation}
  R_1=\cdots=R_G
  \quad\Longrightarrow\quad
  A_1\approx\cdots\approx A_G\approx0.
  \label{eq:saturation}
\end{equation}
Let $g_{\mathcal E}$ and $g_{\mathcal H}$ denote the expected update directions
from easy and hard prompts. Let $\rho_{\mathcal E}$ and $\rho_{\mathcal H}$
denote their sampling weights. The update on the mixed training distribution is
\begin{equation}
  g_{\mathrm{mix}}
  =
  \rho_{\mathcal E}g_{\mathcal E}
  +
  \rho_{\mathcal H}g_{\mathcal H}
  \approx
  \rho_{\mathcal H}g_{\mathcal H}.
  \label{eq:gradient-dominance}
\end{equation}
Thus, hard prompts dominate the update after easy groups become saturated.

However, a zero gradient from easy prompts does not keep their output
distributions fixed. All prompts share the same policy parameters. An update
from hard prompts can therefore change the token probabilities at an easy
prefix $s^{\mathcal E}$. To first order,
\begin{equation}
  \Delta\log\pi_\theta(a\mid s^{\mathcal E})
  \approx
  \eta\rho_{\mathcal H}
  \nabla_\theta\log\pi_\theta(a\mid s^{\mathcal E})^\top
  g_{\mathcal H},
  \label{eq:cross-prompt-drift}
\end{equation}
which is generally nonzero even when $g_{\mathcal E}\approx0$.


\section{Length Self-Distillation}
\label{sec:method}

In this section, we introduce Length Self-Distillation (LSD). It adds two components to the original RL trainer: an
online difficulty router and a temporal self-teacher. 

\begin{figure*}[t]
  \centering
  \includegraphics[width=\textwidth]{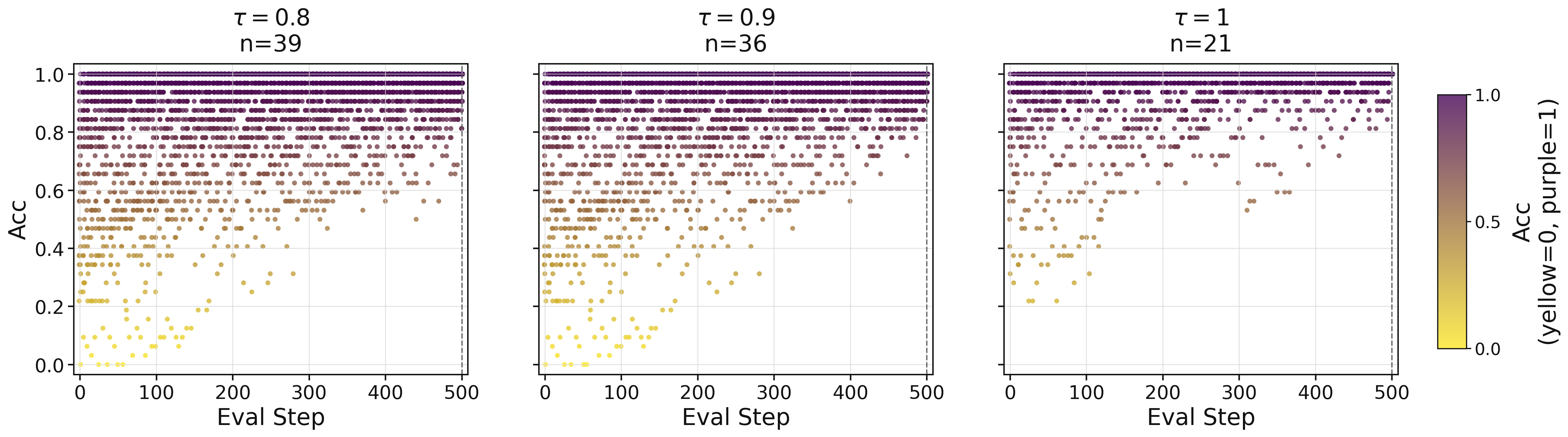}
  \caption{Historical accuracy of prompts selected as easy at step 500. Each point is one prompt at one previous checkpoint. Yellow indicates $A(x)=0$ and purple indicates $A(x)=1$. }
  \label{fig:easy_tracking_history_acc}
\end{figure*}

\paragraph{Online routing.}
Online routing introduces a practical challenge. Ideally, if the prompt in the training batch can be reliably solved by the earlier teacher policy, it should be routed to the OPD objective to preserve the teacher's concise behavior. However, applying this rule directly would require additional teacher rollouts and would roughly double the rollout cost. For efficient training, we use the empirical solve rate of the student’s on-policy rollout group as a lightweight routing signal. To examine the validity of this proxy, we select the easy set at step 500 with thresholds $\tau\in\{0.8,0.9,1.0\}$, and trace the same prompts back through previous checkpoints to check whether earlier policies would also classify them as easy. The result in Figure~\ref{fig:easy_tracking_history_acc} suggests that most of these prompts already have high historical accuracy regardless of the chosen easy-set threshold. Even when using the step-250 checkpoint as the teacher for the step-500 policy, over \(87\%\) of the prompts classified as easy at step-500 are also classified as easy at step-250. 

\paragraph{Temporal self-teacher.}
The simplest self-teacher is the pre-RL policy $\pi_0$. It provides a clean
behavioral anchor because its easy-query responses have not yet accumulated
LST. However, distillation from this fixed teacher can impede further capability acquisition. Since the router uses the current student's solve rate, it may route
newly solved prompts to OPD even when $\pi_0$ cannot solve them reliably.
Figure~\ref{fig:easy_tracking_history_acc} illustrates the underlying historical mismatch. Using \(\pi_0\) as the teacher creates a substantial mismatch between the prompts routed to OPD and those the teacher can solve.
Strong distillation toward this fixed policy can therefore limit benchmark improvement.
 
We address this problem with an \textbf{ exponential moving average (EMA) teacher} $\bar\pi_k$:
\begin{equation}
  \bar\theta_k
  \leftarrow
  \beta\bar\theta_{k-1}
  +(1-\beta)\theta_k,
  \label{eq:ema}
\end{equation}
where $\theta_k$ and $\bar\theta_k$ denote the online policy and the EMA teacher policy's parameters in training step $k$.
The coefficient $\beta\in[0,1]$ controls
the temporal lag of the EMA teacher. A larger $\beta$ keeps the teacher closer to
past policies and provides a stronger behavioral anchor, while a smaller
$\beta$ lets it track the online policy more quickly. In our implementation, we
set $\beta$ through a half-life parameter $H$:
\begin{equation}
  \beta = 2^{-1/H}.
\end{equation}
Thus, the contribution of a past online policy decays exponentially, and its
weight is halved after roughly $H$ EMA updates. By adjusting \(H\), we control how far the teacher lags behind the online policy.

\paragraph{Routed LSD objective.} 
Based on the temporal self-teacher, LSD combines online routing with two different optimization objectives. At training step $k$, the current policy $\pi_k$ generates $G$ responses for each
prompt in the rollout batch $\mathcal B_k$. We compute
$\widehat R(x;\pi_k)$ using these responses and partition the batch into an easy set $\mathcal E_k$ and a hard set $\mathcal H_k$:
\begin{equation}
  \mathcal E_k
  =
  \left\{
    x\in\mathcal B_k:
    \widehat R(x;\pi_k)\ge\tau
  \right\},
  \qquad
  \mathcal H_k
  =
  \mathcal B_k\setminus\mathcal E_k.
  \label{eq:online-routing}
\end{equation}
The original RLVR objective is applied to responses from $\mathcal H_k$.
Responses from $\mathcal E_k$ instead receive an OPD objective defined by the
temporal self-teacher.

We instantiate the OPD objective for easy groups in three ways, yielding three LSD variants that differ in the direction of the Kullback--Leibler (KL) divergence and how its gradient is computed.
\textbf{\emph{Supervised-gradient forward KL (SG-FKL)}} directly minimizes the teacher-to-student KL using the teacher's top-$K$ tokens augmented with stop tokens, with the teacher distribution normalized over this support.
\textbf{\emph{Supervised-gradient reverse KL (SG-RKL)}} instead minimizes the student-to-teacher KL, normalizing both distributions over the same augmented support.
Both supervised-gradient variants differentiate the loss directly through the student logits while treating the teacher distribution as fixed.
\textbf{\emph{Policy-gradient reverse KL (PG-RKL)}} uses the teacher-minus-rollout-policy log-probability difference at each sampled token as a detached advantage in a proximal policy optimization (PPO)-style objective.
Thus, SG-FKL and SG-RKL provide supervision over the full retained support, whereas PG-RKL updates the policy through sampled actions.
All three variants retain the original group-relative RL objective for hard groups. Appendix~\ref{app:objectives} gives the full losses, stop-token treatment, and importance-ratio definitions.

Let $\overline{\ell}^{\mathrm{RL}}$ and $\overline{\ell}^{\mathrm{OPD}}$ be sequence-mean losses on the hard and easy routes, with $n_{\mathcal H}$ and $n_{\mathcal E}$ sampled sequences, respectively.
The number of sequences in each route naturally determines its contribution to the loss. We therefore weight the two route-level losses by their respective numbers of response sequences and get the final objective of LSD:

\begin{equation}
  \mathcal L_{\mathrm{LSD}}
  =
  \frac{n_{\mathcal H}}
       {n_{\mathcal H}+n_{\mathcal E}}
  \overline{\ell}^{\mathrm{RL}}
  +
  \frac{n_{\mathcal E}}
       {n_{\mathcal H}+n_{\mathcal E}}
  \overline{\ell}^{\mathrm{OPD}}.
  \label{eq:lsd-objective}
\end{equation}

\section{Experiments}
\label{sec:experiments}

\subsection{Setup and Comparison Protocol}
\label{sec:setup}

Unless otherwise specified, all main LSD experiments use the training-time routing threshold $\tau=1$. Full configurations are in Appendix~\ref{app:experimental-configurations}.

\paragraph{Single-Turn Reasoning Task.}
We post-train Qwen3-4B-Base on deduplicated DAPO-Math-17K and evaluate AMC 2023 and AIME 2025--2026 with 32 responses per query and a 4k response budget. We compare RL, the three LSD variants, CRISP~\citep{sang2026opsdc}, and Fixed SG-FKL. CRISP distills a periodically refreshed, concise-prompted teacher using reverse KL on all rollouts. Fixed uses a frozen $\pi_0$ teacher, a fixed routing map with threshold 1, and an LSD coefficient of 1. SG-FKL and SG-RKL use $K=32$; EMA updates start after four actor updates. Fixed-easy-set evaluation thresholds are separate from the training routing threshold.

\paragraph{Multi-Turn Agentic Task.}
We follow~\citet{wu2025cutbill} and post-train Qwen3-8B-Base on the CutTheBill training split, evaluating it on BrowseComp-Plus~\citep{chen2025browsecomp}. We use a 20,000-token response budget, at most 48 turns, and a separate Qwen3-8B refinement agent. We compare RL, the three LSD variants, and RL + Length Penalty under the same environment and evaluation configuration. Following AdapThink~\citep{xu2026adapthink}, RL + Length Penalty applies stronger length penalties to easier queries. We report Pass@1, average turns, and $\mathrm{LST}_{50}$. For the agent setting, response length in Eq.~\ref{eq:lst} is the sum of policy-generated tokens across turns; tool observations and refinement-model generation are separate cost components.

\subsection{Main Results}
\label{sec:main-results}
\label{sec:single-turn}

\begin{figure}[ht]
  \centering
  \includegraphics[width=0.90\linewidth]{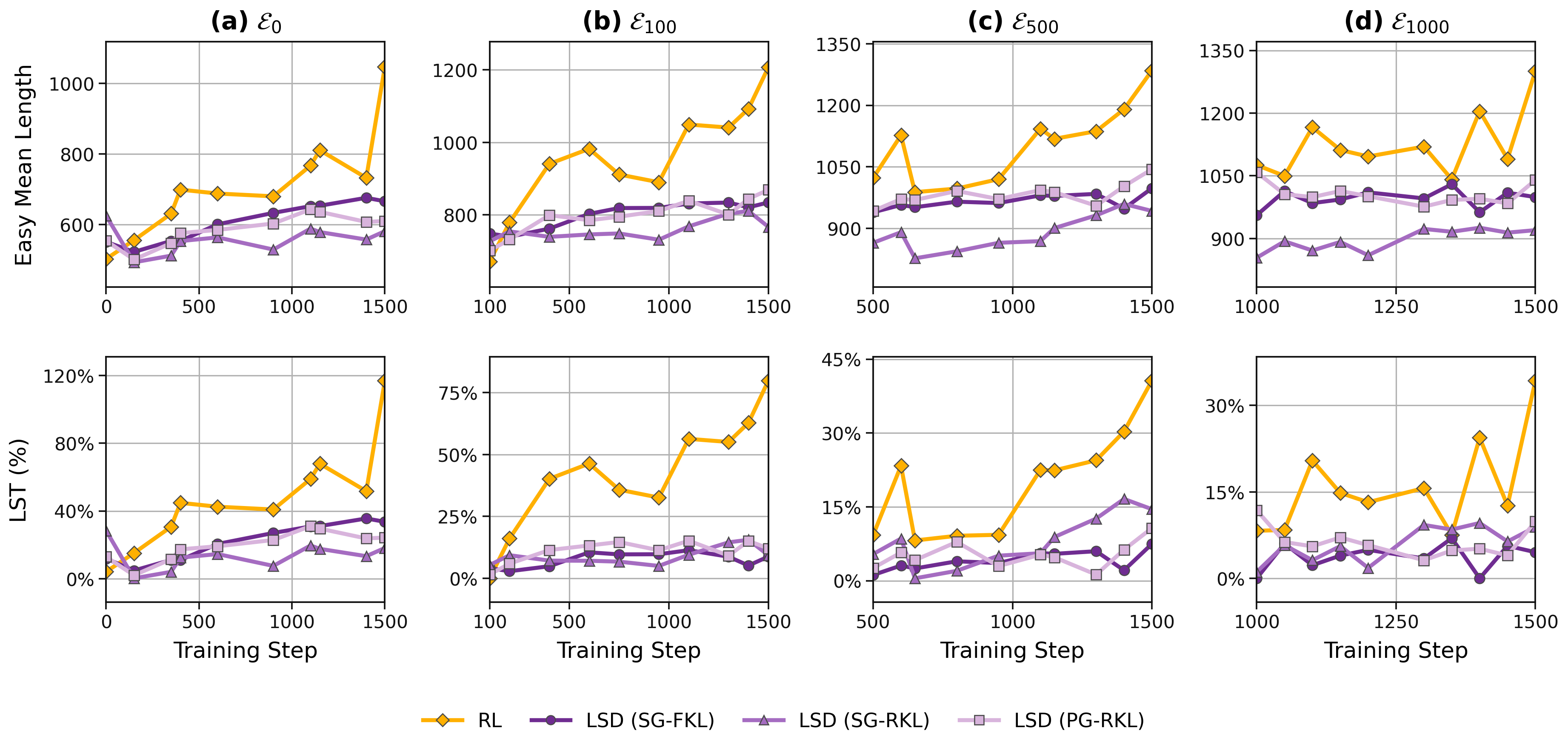}
  \caption{\textbf{Length growth on fixed easy queries.} Across four anchor-defined sets, all three LSD variants have lower easy-query length and LST than RL through most of training. The upper panels report mean length and the lower panels report LST. Different anchors select different cohorts.}
  \label{fig:lsd-easy-tracking}
\end{figure}

\begin{figure}[ht]
  \centering
  \begin{subfigure}[t]{0.49\linewidth}
    \centering
    \includegraphics[width=\linewidth]{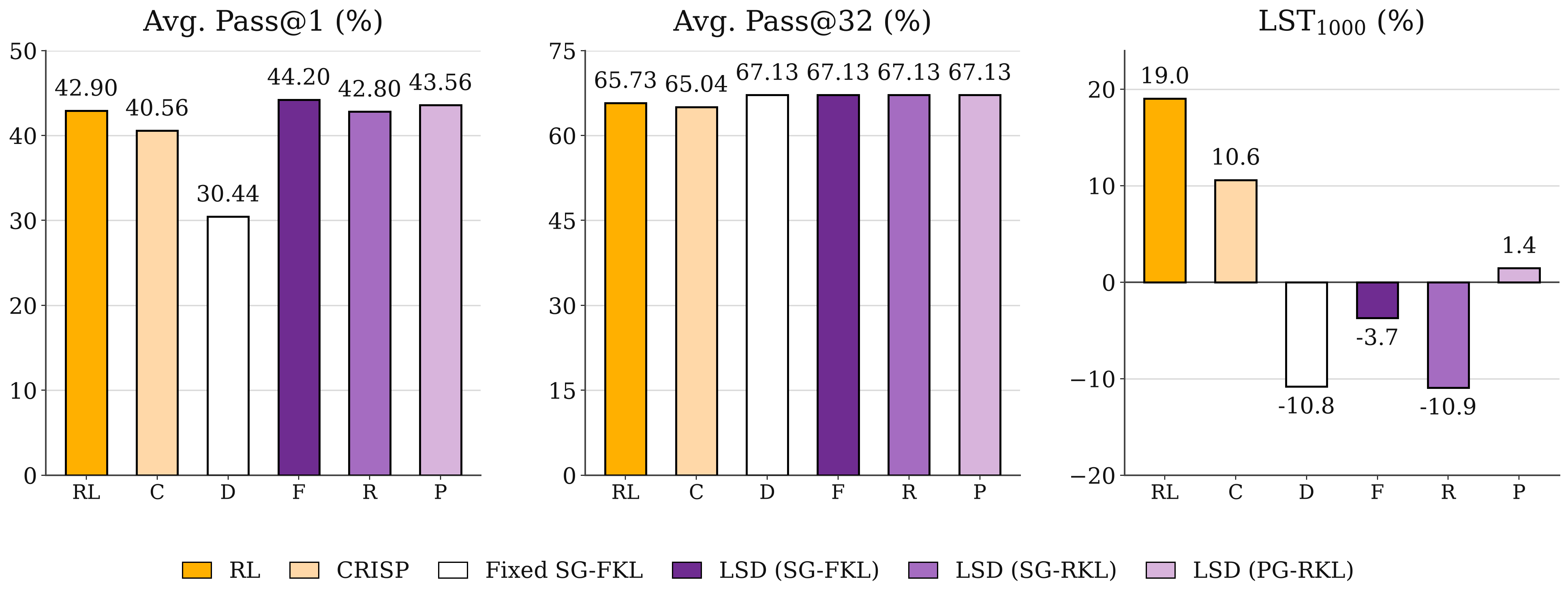}
    \caption{Single-turn reasoning tasks.}
    \label{fig:lsd-selected-summary}
  \end{subfigure}\hfill
  \begin{subfigure}[t]{0.49\linewidth}
    \centering
    \includegraphics[width=\linewidth]{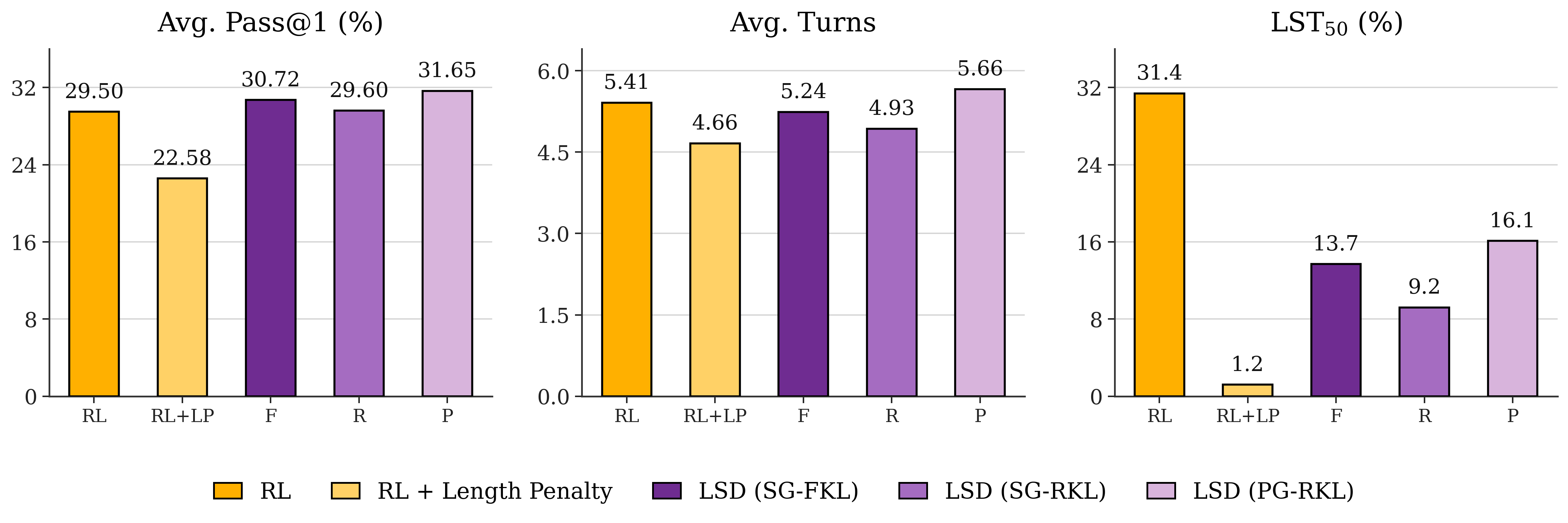}
    \caption{Multi-turn agentic tasks.}
    \label{fig:agent-main}
  \end{subfigure}
  \caption{\textbf{Capability and length-scaling tax across tasks.} (a) Single-turn average Pass@1, Pass@32, and $\mathrm{LST}_{1000}$. (b) BrowseComp-Plus average Pass@1, turns, and $\mathrm{LST}_{50}$. C and D denote CRISP and Fixed SG-FKL; F, R, and P denote LSD with SG-FKL, SG-RKL, and PG-RKL. RL+LP denotes RL + Length Penalty.}
  \label{fig:main-comparison}
\end{figure}

\paragraph{Maintaining concise reasoning on easy queries.}
Figure~\ref{fig:lsd-easy-tracking} shows that LSD curbs easy-query length growth across anchors. Appendix~\ref{app:paired-preservation} jointly reports accuracy and length on identical frozen query sets. Quantitatively, Figure~\ref{fig:lsd-selected-summary} and Table~\ref{tab:lsd-best-each-max} show that, on single-turn reasoning, $\mathrm{LST}_{1000}$ decreases from \mainLSTRLShort\% under RL to \mainLSTFKLShort\%, \mainLSTRKLShort\%, and \mainLSTPGShort\% under SG-FKL, SG-RKL, and PG-RKL, respectively. \label{sec:agent}The same pattern extends to multi-turn agentic tasks. As reported in Figure~\ref{fig:agent-main} and Table~\ref{tab:agent-main-results} in Appendix~\ref{app:benchmark}, $\mathrm{LST}_{50}$ decreases from 31.4\% under RL to 13.7\%, 9.2\%, and 16.1\% under SG-FKL, SG-RKL, and PG-RKL, respectively.

\paragraph{Encouraging more reasoning on hard queries.}
The reduction in easy-query length does not extend to the hard-query cohort. Table~\ref{tab:lsd-best-each-max} reports results on a fixed easy-query set and its hard-query complement. Average hard-query length increases from 2329 tokens under RL to 2511, 2395, and 2393 tokens under SG-FKL, SG-RKL, and PG-RKL, respectively. All three variants therefore produce shorter responses on easy queries while allowing longer responses on hard queries.

The training allocation shows a complementary pattern. As shown in Figure~\ref{fig:training-token-routes}, during training after step 500, the easy route accounts for only 12.17\%, 10.58\%, and 10.94\% of tokens under SG-FKL, SG-RKL, and PG-RKL, respectively. The hard route therefore retains 87.83--89.42\% of the logged token share on average. This allocation keeps training primarily focused on hard queries while preserving concise response patterns on easy queries.

\paragraph{Comparing efficiency baselines.}
CRISP yields 40.56\% Pass@1; Fixed SG-FKL yields 30.44\% Pass@1 and \mainLSTFixedShort\% $\mathrm{LST}_{1000}$ (Figure~\ref{fig:lsd-selected-summary}; Table~\ref{tab:lsd-best-each-max}). On multi-turn tasks, RL + Length Penalty reduces $\mathrm{LST}_{50}$ to 1.2\% and average turns to 4.66, but lowers Pass@1 to 22.58\% (Table~\ref{tab:agent-main-results}).

\paragraph{Comparing the three KL objectives.}
Overall, the three KL objectives differ in how strongly they preserve existing behavior and how they distribute probability across candidate responses. Their Pass@1 scores vary slightly, while their Pass@32 scores are nearly identical. Among the three EMA variants, SG-RKL achieves the lowest LST, but its stronger easy-query compression accompanies lower Pass@1.

The loss definitions offer a possible explanation. SG-FKL weights discrepancies by teacher probabilities, whereas SG-RKL penalizes student mass on tokens assigned low teacher probability. With a lagged teacher, the latter may more strongly preserve established behavior. As shown in Figure~\ref{fig:objective-training-dynamics} and Table~\ref{tab:objective-training-diagnostics}, SG-RKL has lower mean actor entropy (0.0367) than SG-FKL (0.0743) and PG-RKL (0.0542), together with the smallest mean absolute teacher--rollout log-probability difference before updates. Meanwhile, PG-RKL updates sampled actions using the original token probabilities, whereas SG-RKL directly optimizes distributions renormalized on the retained support. This distinction may also contribute to their different outcomes.

\subsection{Ablations}
\label{sec:ablations}

In this section, we ablate two additional hyperparameters introduced by LSD: the EMA half-life $H$ and the routing threshold $\tau$. All ablations use the SG-FKL variant, which achieves the highest average Pass@1 on single-turn reasoning among the three LSD variants.

\begin{figure}[!ht]
  \centering
  \includegraphics[width=0.78\linewidth]{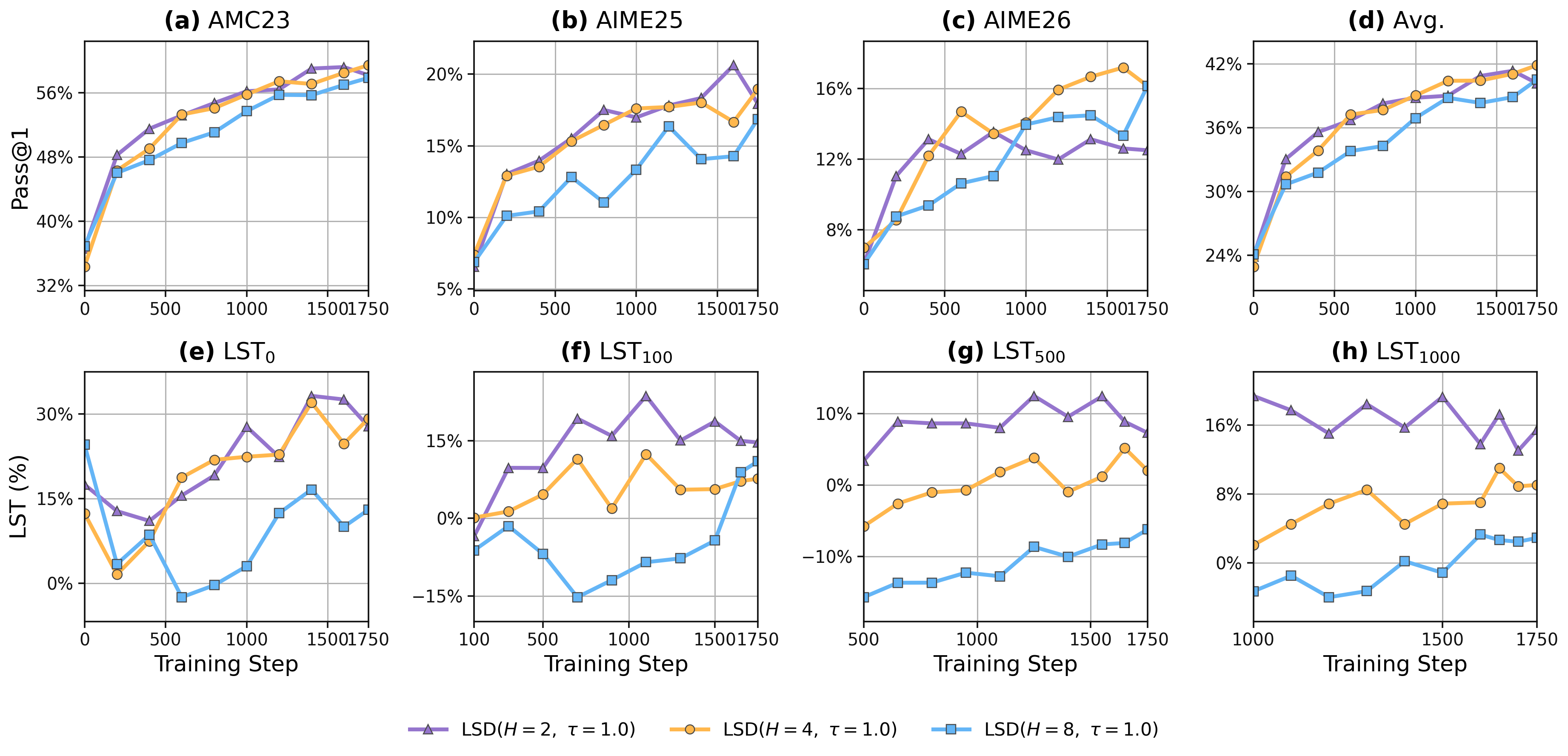}
  \caption{\textbf{EMA half-life ablation.} We compare $H=2,4,8$ at $\tau=1.0$ through step 1750. The upper row shows benchmark and aggregate Pass@1; the lower row shows LST for anchors $b=0,100,500,1000$. }
  \label{fig:ablation-half-life}
\end{figure}

\paragraph{EMA half-life.}
Figure~\ref{fig:ablation-half-life} compares $H\in\{2,4,8\}$ with $\tau=1$. A longer half-life generally suppresses LST but slows capability acquisition. Among the tested settings, $H=4$ achieves the highest average Pass@1 (41.85\%) at step 1750 (Table~\ref{tab:ablation-endpoints}), indicating that keeping the teacher closer to the online policy is not always beneficial. As shown in Figure~\ref{fig:half-life-training-diagnostics} and Table~\ref{tab:ablation-training-diagnostics}, we compare teacher--student parameter lag, distillation loss, and token routing across half-lives. Larger $H$ produces a greater parameter lag and higher distillation loss, while a smaller share of tokens is routed to OPD. The intermediate lag at $H=4$ may provide a stable behavioral reference while allowing the teacher to track improvements in the online policy.

\begin{figure}[!ht]
  \centering
  \includegraphics[width=0.80\linewidth]{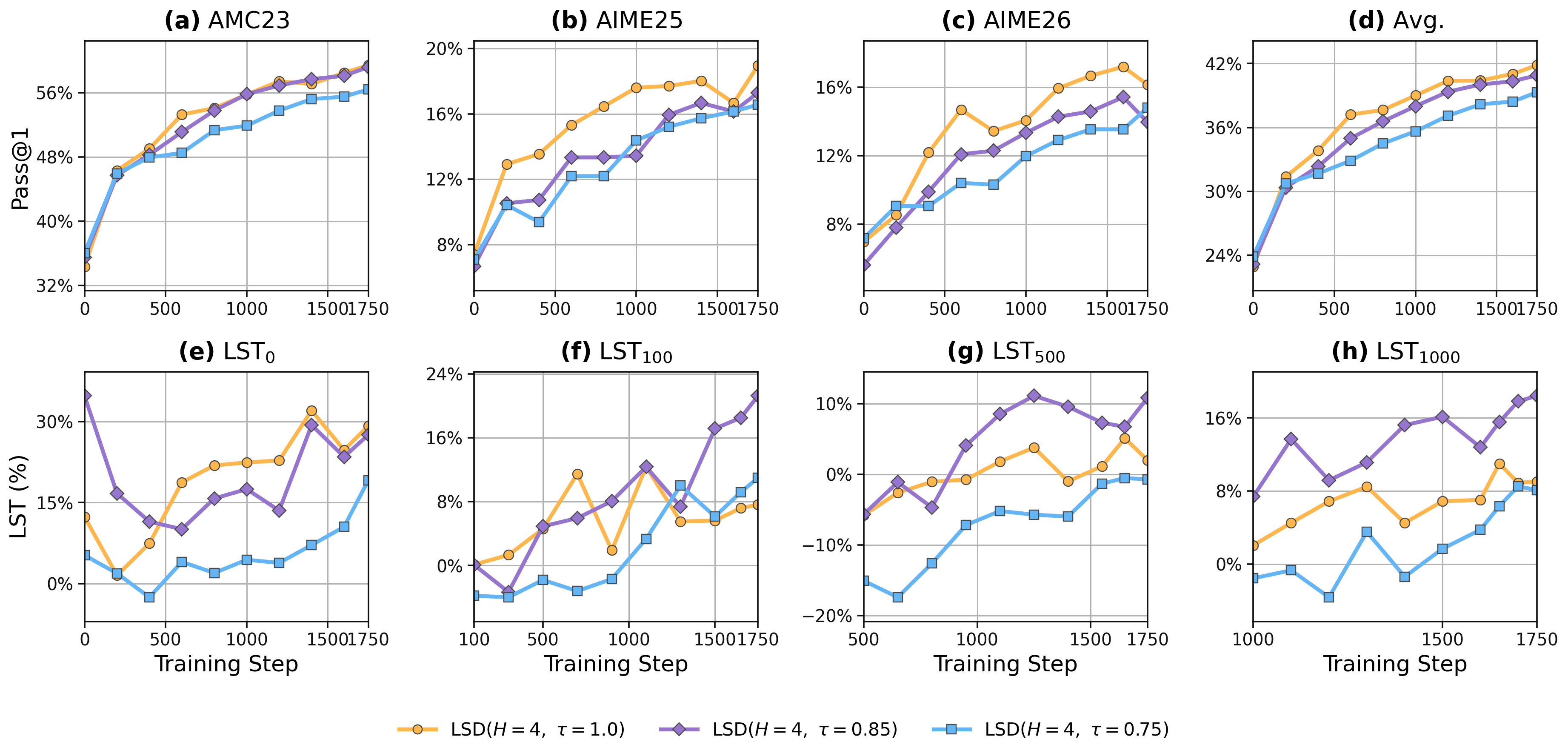}
  \caption{\textbf{Routing-threshold ablation.} We compare $\tau=1.0,0.85,0.75$ at $H=4$ through step 1750. Panels follow Figure~\ref{fig:ablation-half-life}.}
  \label{fig:ablation-threshold}
\end{figure}

\paragraph{Scope of difficulty routing.}
We examine whether preservation should extend to partially solved queries by varying $\tau\in\{1.0,0.85,0.75\}$ with SG-FKL and $H=4$. Lowering $\tau$ routes more partially solved groups to distillation. At step 1750, average Pass@1 decreases from 41.85\% to 40.91\% and 39.31\%, respectively (Figure~\ref{fig:ablation-threshold}; Table~\ref{tab:ablation-endpoints}). Meanwhile, the mean distillation token share increases from 10.31\% to 19.00\% (Figure~\ref{fig:threshold-training-diagnostics}; Table~\ref{tab:ablation-training-diagnostics}). These results support restricting preservation to fully solved rollout groups: such groups lack a group-relative reward signal, whereas partially solved groups still provide reward variation for continued RL learning. Extending preservation before rollout accuracy saturates can therefore compromise capability acquisition. Moreover, to assess the role of query selection, we compare LSD with count-matched random routing under the same EMA and distillation settings (Appendix~\ref{app:paired-preservation}, Table~\ref{tab:paired-main-easy-set}). See Appendix~\ref{app:ablations} for the interpretation of negative LST values.

\section{Conclusion}
In this paper, we identify LST as a conditional cost of RL post-training that arises when responses to already-solved queries lengthen without corresponding accuracy gains. LSD restores a token-level preservation signal on solved rollout groups while retaining RL on unsolved groups, using online accuracy-based routing and an EMA self-teacher. We instantiate LSD with three objectives and demonstrate reduced easy-query cost with strong benchmark performance on single-turn mathematical reasoning and multi-turn agentic tasks. Future work will evaluate LSD on larger models and explore how to extract denser learning signals from sampled data for more effective supervision.
\FloatBarrier
\clearpage
\section*{AI Use Statement}
We used generative AI tools for translation, language editing, literature
synthesis, analysis and plotting code, statistical checks, and discussions
of methodology, experimental design, and result interpretation.
The authors reviewed the AI-assisted outputs and take responsibility for
the final content, including all text, claims, and artifacts.

\section*{Reproducibility Statement}
Training configurations, objectives, and evaluation protocols are documented
in Section~\ref{sec:experiments} and
Appendices~\ref{app:benchmark}--\ref{app:paired-preservation}
and~\ref{app:experimental-configurations}.
We will release our code after paper acceptance.

\bibliographystyle{iclr2027_conference}
\bibliography{main}

@article{wei2022chain,
  title={Chain-of-thought prompting elicits reasoning in large language models},
  author={Wei, Jason and Wang, Xuezhi and Schuurmans, Dale and Bosma, Maarten and Xia, Fei and Chi, Ed and Le, Quoc V. and Zhou, Denny and others},
  journal={Advances in Neural Information Processing Systems},
  volume={35},
  pages={24824--24837},
  year={2022}
}

@article{wang2022self,
  title={Self-consistency improves chain of thought reasoning in language models},
  author={Wang, Xuezhi and Wei, Jason and Schuurmans, Dale and Le, Quoc and Chi, Ed and Narang, Sharan and Chowdhery, Aakanksha and Zhou, Denny},
  journal={arXiv preprint arXiv:2203.11171},
  year={2022}
}

@inproceedings{lightman2024lets,
  title={Let's Verify Step by Step},
  author={Lightman, Hunter and Kosaraju, Vineet and Burda, Yuri and Edwards, Harrison and Baker, Bowen and Lee, Teddy and Leike, Jan and Schulman, John and Sutskever, Ilya and Cobbe, Karl},
  booktitle={International Conference on Learning Representations},
  year={2024}
}

@inproceedings{wanbuffer,
  title={Buffer Matters: Unleashing the Power of Off-Policy Reinforcement Learning in Large Language Model Reasoning},
  author={Wan, Xu and Wang, Yansheng and Huang, Wenqi and Sun, Mingyang},
  booktitle={The Fourteenth International Conference on Learning Representations},
  year={2026}
}

@article{yeo2025demystifying,
  title={Demystifying long chain-of-thought reasoning in llms},
  author={Yeo, Edward and Tong, Yuxuan and Niu, Morry and Neubig, Graham and Yue, Xiang},
  journal={arXiv preprint arXiv:2502.03373},
  year={2025}
}

@article{snell2024scaling,
  title={Scaling LLM Test-Time Compute Optimally Can Be More Effective than Scaling Model Parameters},
  author={Snell, Charlie and Lee, Jaehoon and Xu, Kelvin and Kumar, Aviral},
  journal={arXiv preprint arXiv:2408.03314},
  year={2024}
}

@article{wan2026shadow,
  title={The Shadow Price of Reasoning: Economic Perspective on Optimal Budget Allocation for {LLMs}},
  author={Wan, Xu and Zhu, Speed and Cai, Jianwei and Chen, Guang and Huang, XiMing and Zhou, Wiggin and Sun, Mingyang},
  journal={arXiv preprint arXiv:2606.03092},
  year={2026}
}

@inproceedings{xu2026adapthink,
  title={AdapThink: Adaptive Thinking Preferences for Reasoning Language Models},
  author={Xu, Wenyue and Wan, Xu and Wang, Wei and Huang, Wenqi and Yin, Wotao and Zhao, Shengjie and Sun, Mingyang},
  booktitle={Findings of the Association for Computational Linguistics: ACL 2026},
  pages={9808--9825},
  year={2026}
}

@inproceedings{luo2026o1,
  title={O1-pruner: Length-harmonizing fine-tuning for o1-like reasoning pruning},
  author={Luo, Haotian and He, Haiying and Wang, Yibo and Liu, Shiwei and Li, Wei and Cao, Xiaochun and Tao, Dacheng and Tan, Naiqiang and Shen, Li},
  booktitle={Findings of the Association for Computational Linguistics: ACL 2026},
  pages={14242--14257},
  year={2026}
}

@article{chen2024not,
  title={Do not think that much for 2+ 3=? on the overthinking of o1-like llms},
  author={Chen, Xingyu and Xu, Jiahao and Liang, Tian and He, Zhiwei and Pang, Jianhui and Yu, Dian and Song, Linfeng and Liu, Qiuzhi and Zhou, Mengfei and Zhang, Zhuosheng and others},
  journal={arXiv preprint arXiv:2412.21187},
  year={2024}
}

@article{yang2025qwen3,
  title={Qwen3 technical report},
  author={Yang, An and Li, Anfeng and Yang, Baosong and Zhang, Beichen and Hui, Binyuan and Zheng, Bo and Yu, Bowen and Gao, Chang and Huang, Chengen and Lv, Chenxu and others},
  journal={arXiv preprint arXiv:2505.09388},
  year={2025}
}

@inproceedings{shen2025dast,
  title={Dast: Difficulty-adaptive slow-thinking for large reasoning models},
  author={Shen, Yi and Zhang, Jian and Huang, Jieyun and Shi, Shuming and Zhang, Wenjing and Yan, Jiangze and Wang, Ning and Wang, Kai and Liu, Zhaoxiang and Lian, Shiguo},
  booktitle={Proceedings of the 2025 Conference on Empirical Methods in Natural Language Processing: Industry Track},
  pages={2322--2331},
  year={2025}
}

@article{shao2024deepseekmath,
  title={Deepseekmath: Pushing the limits of mathematical reasoning in open language models},
  author={Shao, Zhihong and Wang, Peiyi and Zhu, Qihao and Xu, Runxin and Song, Junxiao and Bi, Xiao and Zhang, Haowei and Zhang, Mingchuan and Li, YK and Wu, Yang and others},
  journal={arXiv preprint arXiv:2402.03300},
  year={2024}
}

@article{guo2025deepseek,
  title={Deepseek-r1: Incentivizing reasoning capability in llms via reinforcement learning},
  author={Guo, Daya and Yang, Dejian and Zhang, Haowei and Song, Junxiao and Wang, Peiyi and Zhu, Qihao and Xu, Runxin and Zhang, Ruoyu and Ma, Shirong and Bi, Xiao and others},
  journal={arXiv preprint arXiv:2501.12948},
  year={2025}
}

@article{yu2025dapo,
  title={DAPO: An open-source LLM reinforcement learning system at scale},
  author={Yu, Qiying and Zhang, Zheng and Zhu, Ruofei and Yuan, Yufeng and Zuo, Xiaochen and Yue, Yu and Fan, Tiantian and Liu, Gaohong and Liu, Lingjun and Liu, Xin and others},
  journal={arXiv preprint arXiv:2503.14476},
  year={2025}
}

@inproceedings{agarwal2024onpolicy,
  title={On-Policy Distillation of Language Models: Learning from Self-Generated Mistakes},
  author={Agarwal, Rishabh and Vieillard, Nino and Zhou, Yongchao and Stanczyk, Piotr and Garea, Sabela Ramos and Geist, Matthieu and Bachem, Olivier},
  booktitle={International Conference on Learning Representations},
  year={2024}
}

@inproceedings{muennighoff2025s1,
  title={s1: Simple test-time scaling},
  author={Muennighoff, Niklas and Yang, Zitong and Shi, Weijia and Li, Xiang Lisa and Fei-Fei, Li and Hajishirzi, Hannaneh and Zettlemoyer, Luke and Liang, Percy and Cand{\`e}s, Emmanuel and Hashimoto, Tatsunori B},
  booktitle={Proceedings of the 2025 Conference on Empirical Methods in Natural Language Processing},
  pages={20286--20332},
  year={2025}
}

@article{aggarwal2025l1,
  title={L1: Controlling How Long a Reasoning Model Thinks with Reinforcement Learning},
  author={Aggarwal, Pranjal and Welleck, Sean},
  journal={arXiv preprint arXiv:2503.04697},
  year={2025}
}

@article{chen2025browsecomp,
  title={Browsecomp-plus: A more fair and transparent evaluation benchmark of deep-research agent},
  author={Chen, Zijian and Ma, Xueguang and Zhuang, Shengyao and Nie, Ping and Zou, Kai and Liu, Andrew and Green, Joshua and Patel, Kshama and Meng, Ruoxi and Su, Mingyi and others},
  journal={arXiv preprint arXiv:2508.06600},
  year={2025}
}

@misc{wu2025cutbill,
  title        = {Cut the Bill, Keep the Turns: Affordable Multi-Turn Search {RL}},
  author       = {Wu, Jiahao and Xu, Zhongwen and Fu, Qiang and Yang, Wei},
  year         = {2025},
  month        = dec,
  howpublished = {Tencent TEG AIPD Technical Report},
  url          = {https://agate-slipper-ef0.notion.site/Cut-the-Bill-Keep-the-Turns-Affordable-Multi-Turn-Search-RL-003f78214a4d451fb06f453d084e666c},
  note         = {Accessed: 2026-09-04}
}

@article{yu2026dapo,
  title={Dapo: An open-source llm reinforcement learning system at scale},
  author={Yu, Qiying and Zhang, Zheng and Zhu, Ruofei and Yuan, Yufeng and Zuo, Xiaochen and Yue, Yu and Dai, Weinan and Fan, Tiantian and Liu, Gaohong and Liu, Lingjun and others},
  journal={Advances in Neural Information Processing Systems},
  volume={38},
  pages={113222--113244},
  year={2026}
}

@inproceedings{yuan2026shorten,
  title={Shorten after you're right: Lazy length penalties for reasoning rl},
  author={Yuan, Danlong and Xie, Tian and Huang, Shaohan and Zhang, Huishuai and Gong, Zhuocheng and Luo, Chong and Wei, Furu and Zhao, Dongyan},
  booktitle={Findings of the Association for Computational Linguistics: ACL 2026},
  pages={12864--12877},
  year={2026}
}

@article{yi2026shorterbetter,
  title={Shorterbetter: Guiding reasoning models to find optimal inference length for efficient reasoning},
  author={Yi, Jingyang and Wang, Jiazheng and Li, Sida},
  journal={Advances in Neural Information Processing Systems},
  volume={38},
  pages={39011--39043},
  year={2026}
}

@article{xiang2025just,
  title={Just Enough Thinking: Efficient Reasoning with Adaptive Length Penalties Reinforcement Learning},
  author={Xiang, Violet and Blagden, Chase and Rafailov, Rafael and Lile, Nathan and Truong, Sang and Finn, Chelsea and Haber, Nick},
  journal={arXiv preprint arXiv:2506.05256},
  year={2025}
}

@inproceedings{bae2026online,
  title={Online difficulty filtering for reasoning oriented reinforcement learning},
  author={Bae, Sanghwan and Hong, Jiwoo and Lee, Min Young and Kim, Hanbyul and Nam, JeongYeon and Kwak, Donghyun},
  booktitle={Proceedings of the 19th Conference of the European Chapter of the Association for Computational Linguistics (Volume 1: Long Papers)},
  pages={700--719},
  year={2026}
}

@misc{openai2025o3,
  author={{OpenAI}},
  title={Introducing OpenAI o3 and o4-mini},
  year={2025},
  howpublished={\url{https://openai.com/index/introducing-o3-and-o4-mini/}}
}

@misc{anthropic2025thinking,
  author={{Anthropic}},
  title={Claude's Extended Thinking},
  year={2025},
  howpublished={\url{https://www.anthropic.com/news/visible-extended-thinking}}
}

@misc{google2025gemini,
  author={{Google}},
  title={Gemini 2.5 Thinking Model Updates},
  year={2025},
  howpublished={\url{https://developers.googleblog.com/gemini-2-5-thinking-model-updates/}}
}

@article{sang2026opsdc,
  title={CRISP: Compressed Reasoning via Iterative Self-Policy Distillation},
  author={Sang, Hejian and Xu, Yuanda and Zhou, Zhengze and He, Ran and Wang, Zhipeng and Sun, Jiachen},
  journal={arXiv preprint arXiv:2603.05433},
  year={2026}
}

@article{qu2026gps,
  title={Small Generalizable Prompt Predictive Models Can Steer Efficient RL Post-Training of Large Reasoning Models},
  author={Qu, Yun and Wang, Qi and Mao, Yixiu and Zou, Heming and Jiang, Yuhang and Liu, Weijie and Bai, Clive and Yang, Kai and Chen, Yangkun and Yang, Saiyong and Ji, Xiangyang},
  journal={arXiv preprint arXiv:2602.01970},
  year={2026}
}

@article{jin2025search,
  title={Search-r1: Training llms to reason and leverage search engines with reinforcement learning},
  author={Jin, Bowen and Zeng, Hansi and Yue, Zhenrui and Yoon, Jinsung and Arik, Sercan and Wang, Dong and Zamani, Hamed and Han, Jiawei},
  journal={arXiv preprint arXiv:2503.09516},
  year={2025}
}

@article{ruan2026contrastive,
  title={Contrastive On-Policy Distillation},
  author={Ruan, Jiacheng and Tang, Jun and Yuan, Wenzhen and Liu, Ting and Bai, Shuai and Liu, Dayiheng and Yang, Zhibo and Fu, Yuzhuo},
  journal={arXiv preprint arXiv:2607.19046},
  year={2026}
}

\clearpage
\appendix
\section{Additional benchmark results}
\label{app:benchmark}

\subsection{Single-turn reasoning}
\begin{table}[htbp]
  \centering
  \caption{\textbf{Evaluation results for single-turn reasoning.}
  Panel (a) reports accuracy in percent; panel (b) reports the mean token counts, and LST.
  Avg.\ Acc.\ weights benchmarks by their numbers of evaluation prompts.
  Avg.\ Len.\ covers all 143 prompts. Easy and Hard Len.\ use the fixed
  RL step-100 easy set ($A(x)\ge 0.9$) and its hard-query complement.
  $\mathrm{LST}_{1000}$ uses the $\tau=1$ RL step-1000 cohort in
  Table~\ref{tab:paired-preservation}. On this fixed easy set, RL step
  \sharedLSTReferenceStep{} has the lowest mean response length among RL
  checkpoints satisfying the accuracy constraint (100\%). Its mean length,
  \sharedLSTReferenceLength{} tokens, is the shared reference for all methods.}
  \label{tab:lsd-best-each-max}
  \small
  \resizebox{\textwidth}{!}{%
  \begin{tabular}{lrrrrrrrr}
    \toprule
    \multicolumn{9}{l}{\textbf{(a) Task performance (\%)}} \\
    \midrule
    & \multicolumn{2}{c}{\textbf{AMC}}
    & \multicolumn{2}{c}{\textbf{AIME25}}
    & \multicolumn{2}{c}{\textbf{AIME26}}
    & \multicolumn{2}{c}{\textbf{Avg. Acc.}} \\
    \cmidrule(lr){2-3}
    \cmidrule(lr){4-5}
    \cmidrule(lr){6-7}
    \cmidrule(lr){8-9}
    Method
    & Pass@1 & Pass@32
    & Pass@1 & Pass@32
    & Pass@1 & Pass@32
    & Pass@1 & Pass@32 \\
    \midrule
    \methodname{RL}{RL}
      & 60.09 & 84.34
      & \underline{22.19} & \textbf{43.33}
      & 16.04 & \underline{36.67}
      & 42.90 & \underline{65.73} \\

    \methodname{CRISP}{CRISP}
      & 58.48 & \underline{86.75}
      & 18.36 & 36.67
      & 13.18 & 33.33
      & 40.56 & 65.04 \\

    \methodname{Fixed}{Fixed SG-FKL}
      & 45.07 & \textbf{87.95}
      & 11.46 & \underline{40.00}
      & 8.96 & \underline{36.67}
      & 30.44 & \textbf{67.13} \\

    \methodname{SGFKL}{LSD (SG-FKL)}
      & \textbf{61.30} & \underline{86.75}
      & \textbf{22.92} & \textbf{43.33}
      & \textbf{18.15} & \underline{36.67}
      & \textbf{44.20} & \textbf{67.13} \\

    \methodname{SGRKL}{LSD (SG-RKL)}
      & \underline{60.87} & \textbf{87.95}
      & 19.65 & 36.67
      & 15.94 & \textbf{40.00}
      & 42.80 & \textbf{67.13} \\

    \methodname{PGRKL}{LSD (PG-RKL)}
      & 60.70 & \textbf{87.95}
      & 21.96 & \textbf{43.33}
      & \underline{17.76} & 33.33
      & \underline{43.56} & \textbf{67.13} \\
    \bottomrule
  \end{tabular}
  }
  \par\medskip
  \resizebox{\textwidth}{!}{%
  \begin{tabular}{lrrrrr}
    \toprule
    \multicolumn{6}{l}{\textbf{(b) Response length and LST}} \\
    \midrule
    Method & Avg.\ Len. & Easy Len. & Hard Len.
      & $\mathrm{LST}_{0}$ (\%) & $\mathrm{LST}_{1000}$ (\%) \\
    \midrule
    \methodname{RL}{RL}                 & \underline{2124} & 997 & \underline{2329} & 67.2 & \mainLSTRL \\
    \methodname{CRISP}{CRISP}              & 2150 & 972 & 2364 & 61.2 & \mainLSTCRISP \\
    \methodname{Fixed}{Fixed SG-FKL}   & \textbf{1226} & \textbf{691} & \textbf{1323} & \underline{11.0} & \underline{\mainLSTFixed} \\
    \methodname{SGFKL}{LSD (SG-FKL)} & 2257 & 859 & 2511 & 31.5 & \mainLSTFKL \\
    \methodname{SGRKL}{LSD (SG-RKL)} & 2143 & \underline{757} & 2395 & \textbf{9.1} & $\mathbf{\mainLSTRKL}$ \\
    \methodname{PGRKL}{LSD (PG-RKL)} & 2148 & 802 & 2393 & 14.0 & \mainLSTPG \\
    \bottomrule
  \end{tabular}%
  }
\end{table}

The fixed hard-query cohort is the complement
of the evaluation easy set; it is distinct from the dynamically routed
hard groups used during training.

\subsection{Multi-turn agentic tasks}
\begin{table}[htbp]
  \centering
  \caption{\textbf{Evaluation results on multi-turn agentic tasks.}
  Values correspond to Figure~\ref{fig:agent-main}.
  Pass@1 and LST are in percent; lengths are in tokens.}
  \label{tab:agent-main-results}
  \small
  \setlength{\tabcolsep}{5pt}
  \resizebox{\linewidth}{!}{%
  \begin{tabular}{lrrrrrrr}
    \toprule
    \multirow{2}{*}{Method} & \multicolumn{7}{c}{\textbf{BrowseComp-Plus}} \\
    \cmidrule(lr){2-8}
    & Pass@1 & Avg.\ Turns & Easy Len. & Hard Len.
      & $\mathrm{LST}_{0}$ & $\mathrm{LST}_{50}$ & $\mathrm{LST}_{100}$ \\
    \midrule
    \methodname{RL}{RL}                 & 29.50 & 5.41 & 3531.9 & 4065.3 & 42.3 & 31.4 & 11.8 \\
    \methodname{LP}{RL + Length Penalty} & 22.58 & \textbf{4.66} & 2708 & 3426 & & \textbf{1.2} & \\
    \methodname{SGFKL}{LSD (SG-FKL)} & \underline{30.72} & 5.24 & \underline{2706.0} & 4252.0 & \underline{24.9} & 13.7 & \underline{1.2} \\
    \methodname{SGRKL}{LSD (SG-RKL)} & 29.60 & \underline{4.93} & \textbf{2146.5} & 4034.8 & \textbf{13.6} & \underline{9.2} & $\mathbf{-6.2}$ \\
    \methodname{PGRKL}{LSD (PG-RKL)} & \textbf{31.65} & 5.66 & 2715.2 & 4139.5 & 31.4 & 16.1 & 2.4 \\
    \bottomrule
  \end{tabular}%
  }
\end{table}

SG-FKL and SG-RKL reduce both LST and average interaction turns while increasing
Pass@1. PG-RKL attains the highest Pass@1 but uses more turns than RL,
showing that lower LST does not necessarily imply fewer interactions.
RL + Length Penalty obtains the lowest $\mathrm{LST}_{50}$ and fewest turns, but its
Pass@1 falls to 22.58\% from RL's 29.50\%.


\clearpage
\section{Ablation reporting details}
\label{app:ablations}

The ablation trajectories are shown in Figures~\ref{fig:ablation-half-life} and~\ref{fig:ablation-threshold} in Section~\ref{sec:ablations}.

The ablations use a fixed, accuracy-qualified reference length from aligned RL, shared across configurations. Negative LST indicates responses shorter than this reference; accuracy preservation is evaluated separately on the same fixed easy set.

\begin{table}[htbp]
  \centering
  \caption{\textbf{Ablation endpoint summary at step 1750.}
  Avg.\ Pass@1 is weighted over all 143 evaluation prompts.
  LST is computed against the fixed aligned-RL accuracy-qualified reference.
  The $H=4,\tau=1.0$ run is shared by both comparisons.}
  \label{tab:ablation-endpoints}
  \small
  \resizebox{\linewidth}{!}{%
  \begin{tabular}{ccrrrrr}
    \toprule
    $H$ & $\tau$ & Avg.\ Pass@1 (\%)
    & $\mathrm{LST}_0$ (\%) & $\mathrm{LST}_{100}$ (\%)
    & $\mathrm{LST}_{500}$ (\%) & $\mathrm{LST}_{1000}$ (\%) \\
    \midrule
    2 & 1.00 & 40.14 & 27.79 & 14.64 &  7.29 & 15.49 \\
    4 & 1.00 & 41.85 & 29.20 &  7.63 &  2.01 &  9.06 \\
    8 & 1.00 & 40.49 & 13.01 & 11.00 & -6.21 &  2.91 \\
    4 & 0.85 & 40.91 & 27.55 & 21.24 & 10.84 & 18.42 \\
    4 & 0.75 & 39.31 & 19.15 & 10.97 & -0.68 &  8.11 \\
    \bottomrule
  \end{tabular}%
  }
\end{table}

We compare all five configurations over the shared steps 1--1750.
Table~\ref{tab:ablation-training-diagnostics} reports arithmetic means
of logged per-step scalars; the Pass@1 column instead gives the
unsmoothed evaluation at step 1750. The fresh $H=4,\tau=1.0$ run is
used in both ablations.

\begin{table}[htbp]
  \centering
  \caption{\textbf{LSD training signals in the $H$ and $\tau$ ablations.}
  Sequence and token shares refer to the easy route. Parameter gap is
  the logged maximum absolute teacher--student parameter difference
  before the EMA update. Training statistics average 1750 steps per run.}
  \label{tab:ablation-training-diagnostics}
  \small
  \resizebox{\linewidth}{!}{%
  \begin{tabular}{ccrrrrrr}
    \toprule
    $H$ & $\tau$ & \shortstack{Pass@1\\(\%)}
      & \shortstack{Easy seq.\\(\%)} & \shortstack{Easy tok.\\(\%)}
      & \shortstack{Effective\\OPD coef.}
      & \shortstack{Param. gap\\($\times10^{-5}$)}
      & \shortstack{OPD loss\\($\times10^{-3}$)} \\
    \midrule
    2 & 1.00 & 40.14 & 18.87 & 10.79 & 0.252 & 0.690 & 0.614 \\
    4 & 1.00 & 41.85 & 18.54 & 10.31 & 0.247 & 1.087 & 0.693 \\
    8 & 1.00 & 40.49 & 15.24 &  7.90 & 0.192 & 1.690 & 0.841 \\
    4 & 0.85 & 40.91 & 24.67 & 14.87 & 0.354 & 1.077 & 0.715 \\
    4 & 0.75 & 39.31 & 29.50 & 19.00 & 0.450 & 1.078 & 0.689 \\
    \bottomrule
  \end{tabular}%
  }
\end{table}

\clearpage
\subsection{Teacher lag and distillation exposure}
\label{app:ablation-training-diagnostics}

Increasing $H$ produces a larger teacher parameter lag and a larger
logged SG-FKL loss, while reducing the fraction of tokens sent to OPD.
Thus, the stronger preservation observed for $H=8$ does not require
greater distillation exposure. Instead, the delayed teacher can impose
a stronger constraint on the tokens it supervises. The maximum
parameter gap is a parameter-space diagnostic, not a KL divergence;
its ordering need not match the teacher--rollout log-probability gap.

\begin{figure}[htbp]
  \centering
  \includegraphics[width=0.90\linewidth]{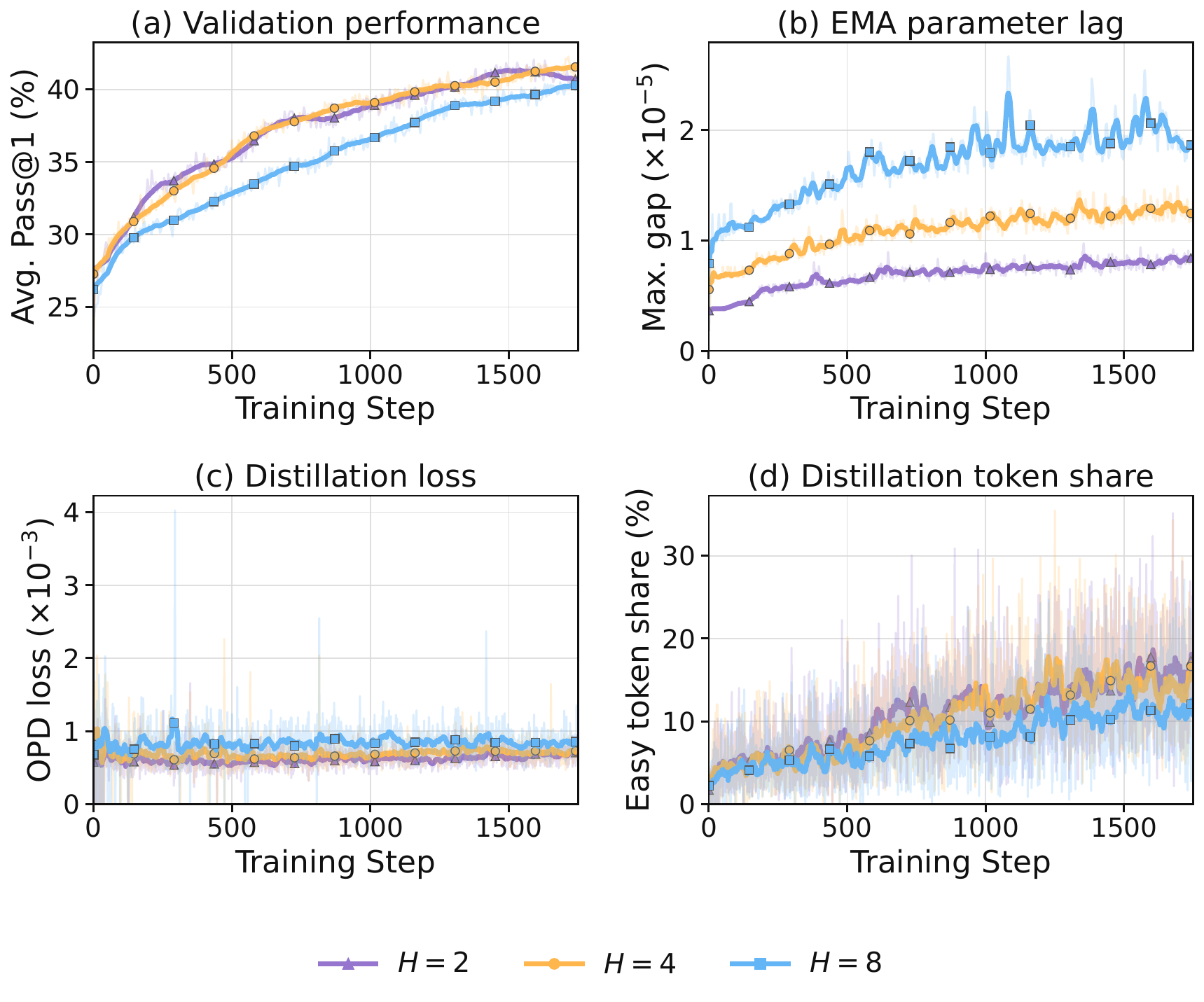}
  \caption{\textbf{Training signals behind the EMA half-life ablation.}
  All runs use SG-FKL and $\tau=1$.
  Faint lines show raw values; marked curves show centered 21-point
  moving means, using available points at the boundaries.
  Evaluation points are five training steps apart; the other metrics
  are logged every training step.}
  \label{fig:half-life-training-diagnostics}
\end{figure}

\clearpage
\subsection{Routing threshold and preservation pressure}

\begin{figure}[htbp]
  \centering
  \includegraphics[width=0.90\linewidth]{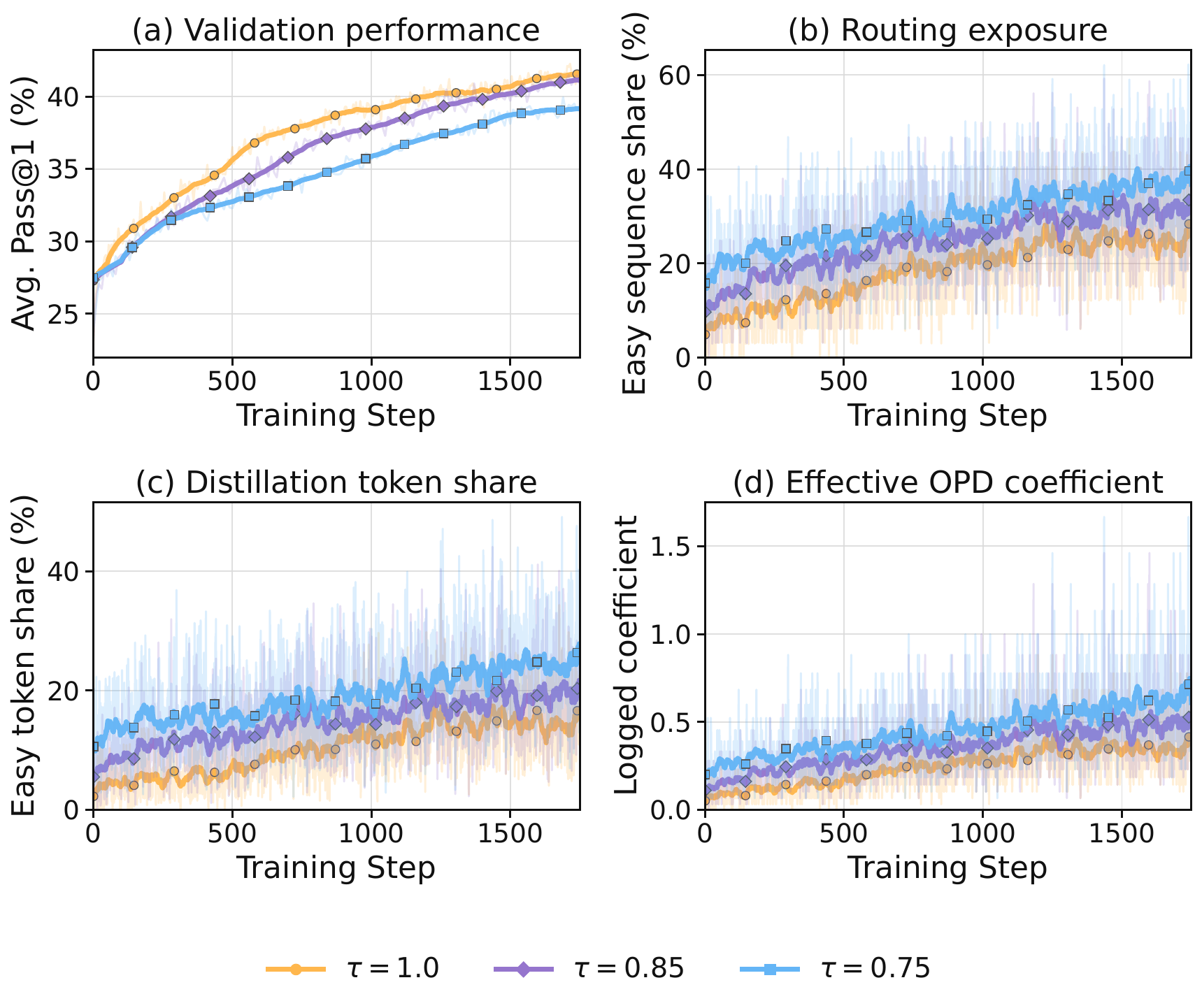}
  \caption{\textbf{Training signals behind the routing-threshold ablation.}
  All runs use SG-FKL and $H=4$, over steps 1--1750.
  Raw and smoothed curves follow the convention in
  Figure~\ref{fig:half-life-training-diagnostics}.
  Lower thresholds increase the number of sequences and fraction of
  tokens routed to distillation, as well as the effective OPD coefficient.}
  \label{fig:threshold-training-diagnostics}
\end{figure}

With eight responses per group, thresholds $\tau=1$, $0.85$, and $0.75$
admit groups with at least eight, seven, and six correct responses,
respectively. Lowering the threshold therefore moves more partially
solved groups to distillation. As shown in
Figure~\ref{fig:threshold-training-diagnostics}, both the easy-route
sequence share and token share increase, along with the logged
effective OPD coefficient. From $\tau=1$ to $0.75$, the mean token
share rises from 10.31\% to 19.00\%, while the effective coefficient
rises from 0.247 to 0.450. The lower Pass@1 at step 1750 is consistent
with more preservation pressure on prompts that still admit incorrect
responses. These coupled changes do not isolate which training signal
causes the performance difference.

\FloatBarrier
\section{Detailed RL and distillation objectives}
\label{app:objectives}

The notation and losses below expand the routed objective in Section~\ref{sec:method}.

For each hard prompt $x$, let $\{y_j\}_{j=1}^{G}$ denote its rollout group and
let $r_j=r(x,y_j)$ be the reward of response $y_j$. We compute the
group-relative advantage as
\begin{equation}
  A_i
  =
  \frac{r_i-\overline r_x}{\sigma_x+\epsilon_{\mathrm{adv}}},
  \qquad
  \overline r_x
  =
  \frac{1}{G}\sum_{j=1}^{G}r_j,
  \qquad
  \sigma_x
  =
  \sqrt{
    \frac{1}{G}
    \sum_{j=1}^{G}
    (r_j-\overline r_x)^2
  }.
  \label{eq:group-relative-advantage}
\end{equation}
For a response $y_i$ of length $T_i$, let
$s_{i,t}=(x,y_{i,<t})$ denote its prefix at token $t$. The importance ratio is
\begin{equation}
  \rho_{i,t}(\theta)
  =
  \frac{\pi_\theta(y_{i,t}\mid s_{i,t})}
       {\pi_{\mathrm{old}}(y_{i,t}\mid s_{i,t})}.
\end{equation}
Its sequence-level RL loss is
\begin{equation}
  \ell_i^{\mathrm{RL}}
  =
  -\frac{1}{T_i}
  \sum_{t=1}^{T_i}
  \min\left(
    \rho_{i,t}(\theta)A_i,\,
    \operatorname{clip}
    \bigl(\rho_{i,t}(\theta),1-\epsilon,1+\epsilon\bigr)A_i
  \right).
  \label{eq:rl-sequence-loss}
\end{equation}

For an easy response $y_i$, we consider three OPD variants. Two directly
optimize a top-$K$ KL objective, while the third estimates reverse KL through
policy gradients.

\paragraph{Supervised-gradient forward KL (SG-FKL).}
Let $\mathcal V_{i,t}^{K}$ denote the teacher's top-$K$ tokens at prefix
$s_{i,t}$. To ensure that termination behavior is supervised even when a stop token does not appear in the teacher's top-$K$ predictions, we augment this set with the stop-token set
$\mathcal V_{\mathrm{stop}}$:
\begin{equation}
  \widetilde{\mathcal V}_{i,t}^{K}
  =
  \mathcal V_{i,t}^{K}
  \cup
  \mathcal V_{\mathrm{stop}}.
\end{equation}

We normalize the teacher distribution over the augmented support:
\begin{equation}
  \bar\pi_k^{K}(v\mid s_{i,t})
  =
  \frac{
    \bar\pi_k(v\mid s_{i,t})
  }{
    \sum_{u\in\widetilde{\mathcal V}_{i,t}^{K}}
    \bar\pi_k(u\mid s_{i,t})
  },
  \qquad
  v\in\widetilde{\mathcal V}_{i,t}^{K}.
\end{equation}
Then, the forward-KL OPD objective is
\begin{equation}
  \ell_i^{\mathrm{SG\text{-}FKL}}
  =
  \frac{1}{T_i}
  \sum_{t=1}^{T_i}
  \sum_{v\in\widetilde{\mathcal V}_{i,t}^{K}}
  \bar\pi_k^{K}(v\mid s_{i,t})
  \left[
    \log\bar\pi_k^{K}(v\mid s_{i,t})
    -
    \log\pi_\theta(v\mid s_{i,t})
  \right].
  \label{eq:opd-sg-fkl}
\end{equation}
Gradients are taken directly through the student logits, while the teacher
distribution is detached.

\paragraph{Supervised-gradient reverse KL (SG-RKL).}
For the reverse direction, we also normalize the student distribution over
the same augmented support:
\begin{equation}
  \pi_\theta^{K}(v\mid s_{i,t})
  =
  \frac{
    \pi_\theta(v\mid s_{i,t})
  }{
    \sum_{u\in\widetilde{\mathcal V}_{i,t}^{K}}
    \pi_\theta(u\mid s_{i,t})
  },
  \qquad
  v\in\widetilde{\mathcal V}_{i,t}^{K}.
\end{equation}
We then directly minimize
\begin{equation}
  \ell_i^{\mathrm{SG\text{-}RKL}}
  =
  \frac{1}{T_i}
  \sum_{t=1}^{T_i}
  \sum_{v\in\widetilde{\mathcal V}_{i,t}^{K}}
  \pi_\theta^{K}(v\mid s_{i,t})
  \left[
    \log\pi_\theta^{K}(v\mid s_{i,t})
    -
    \log\bar\pi_k^{K}(v\mid s_{i,t})
  \right].
  \label{eq:opd-sg-rkl}
\end{equation}

\paragraph{Policy-gradient reverse KL (PG-RKL).}
The third objective only uses the teacher probability of the sampled token.
For each token, we define the detached OPD advantage
\begin{equation}
  A_{i,t}^{\mathrm{OPD}}
  =
  \operatorname{sg}\left[
    \log\bar\pi_k(y_{i,t}\mid s_{i,t})
    -
    \log\pi_{\mathrm{old}}(y_{i,t}\mid s_{i,t})
  \right],
  \label{eq:opd-advantage}
\end{equation}
where $\operatorname{sg}[\cdot]$ denotes stop-gradient. We then apply the
PPO-style clipped objective:
\begin{equation}
  \ell_i^{\mathrm{PG\text{-}RKL}}
  =
  -\frac{1}{T_i}
  \sum_{t=1}^{T_i}
  \min\left(
    \rho_{i,t}(\theta)A_{i,t}^{\mathrm{OPD}},
    \operatorname{clip}
    \bigl(\rho_{i,t}(\theta),1-\epsilon,1+\epsilon\bigr)
    A_{i,t}^{\mathrm{OPD}}
  \right).
  \label{eq:opd-pg-rkl}
\end{equation}

Let $\ell_i^{\mathrm{OPD}}$ denote the OPD loss used by a particular LSD
variant. Under sequence-mean--token-mean aggregation, we first average the
token losses within each response. We then average the resulting
sequence-level losses within their assigned routes:
\begin{equation}
  \overline{\ell}^{\mathrm{RL}}
  =
  \frac{1}{n_{\mathcal H}}
  \sum_{i\in\mathcal Y_{\mathcal H}}
  \ell_i^{\mathrm{RL}},
  \qquad
  \overline{\ell}^{\mathrm{OPD}}
  =
  \frac{1}{n_{\mathcal E}}
  \sum_{i\in\mathcal Y_{\mathcal E}}
  \ell_i^{\mathrm{OPD}},
  \label{eq:route-losses}
\end{equation}
where $\mathcal Y_{\mathcal H}$ and $\mathcal Y_{\mathcal E}$ contain the
response sequences routed to RLVR and OPD, respectively, and
$n_{\mathcal H}=|\mathcal Y_{\mathcal H}|$ and
$n_{\mathcal E}=|\mathcal Y_{\mathcal E}|$.

\clearpage
\subsection{Logged optimization diagnostics}
\label{app:training-diagnostics}

We examine the logged training histories of the three single-turn LSD
runs. The shared training
interval is steps 501--1628: SG-RKL's available training history begins
at step 501, and PG-RKL's ends at step 1628. All three configurations
record an EMA half-life of four updates, an EMA update interval of one,
and an online routing threshold of one.

\begin{figure}[htbp]
  \centering
  \includegraphics[width=\linewidth]{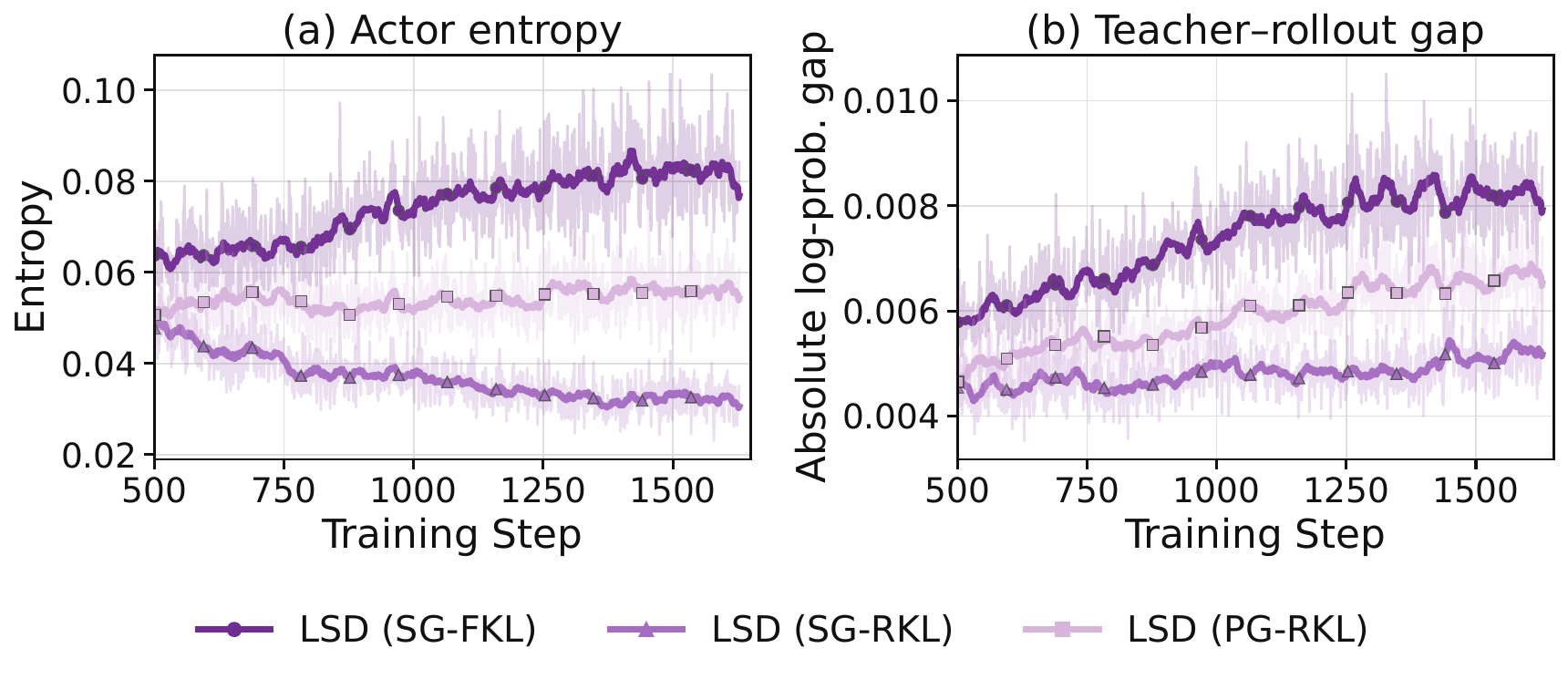}
  \caption{\textbf{Training dynamics of the three LSD objectives.}
  Actor entropy (left) and the mean absolute teacher--rollout
  log-probability difference before updates (right) over the shared
  steps 501--1628. Faint traces show raw per-step values; marked curves
  show centered 21-step means, using available points at the boundaries. SG-RKL exhibits
  lower entropy and a smaller teacher--rollout discrepancy over this
  interval. Each trajectory uses its own run's training batches.}
  \label{fig:objective-training-dynamics}
\end{figure}

\begin{table}[htbp]
  \centering
  \caption{\textbf{Optimization diagnostics over a shared training interval.}
  Values are arithmetic means of the logged per-step scalars over all
  1128 shared steps, without smoothing. These training steps are not
  independent experimental replicates.}
  \label{tab:objective-training-diagnostics}
  \small
  \begin{tabular}{lrrr}
    \toprule
    Variant & Actor entropy & Teacher--rollout gap & Easy sequences (\%) \\
    \midrule
    \methodname{SGFKL}{SG-FKL} & 0.0743 & 0.00738 & 21.51 \\
    \methodname{SGRKL}{SG-RKL} & 0.0367 & 0.00481 & 20.23 \\
    \methodname{PGRKL}{PG-RKL} & 0.0542 & 0.00590 & 20.67 \\
    \bottomrule
  \end{tabular}
\end{table}

\clearpage
\subsection{Easy and hard token allocation during training}
\label{app:training-token-routes}

\begin{figure}[htbp]
  \centering
  \includegraphics[width=\linewidth]{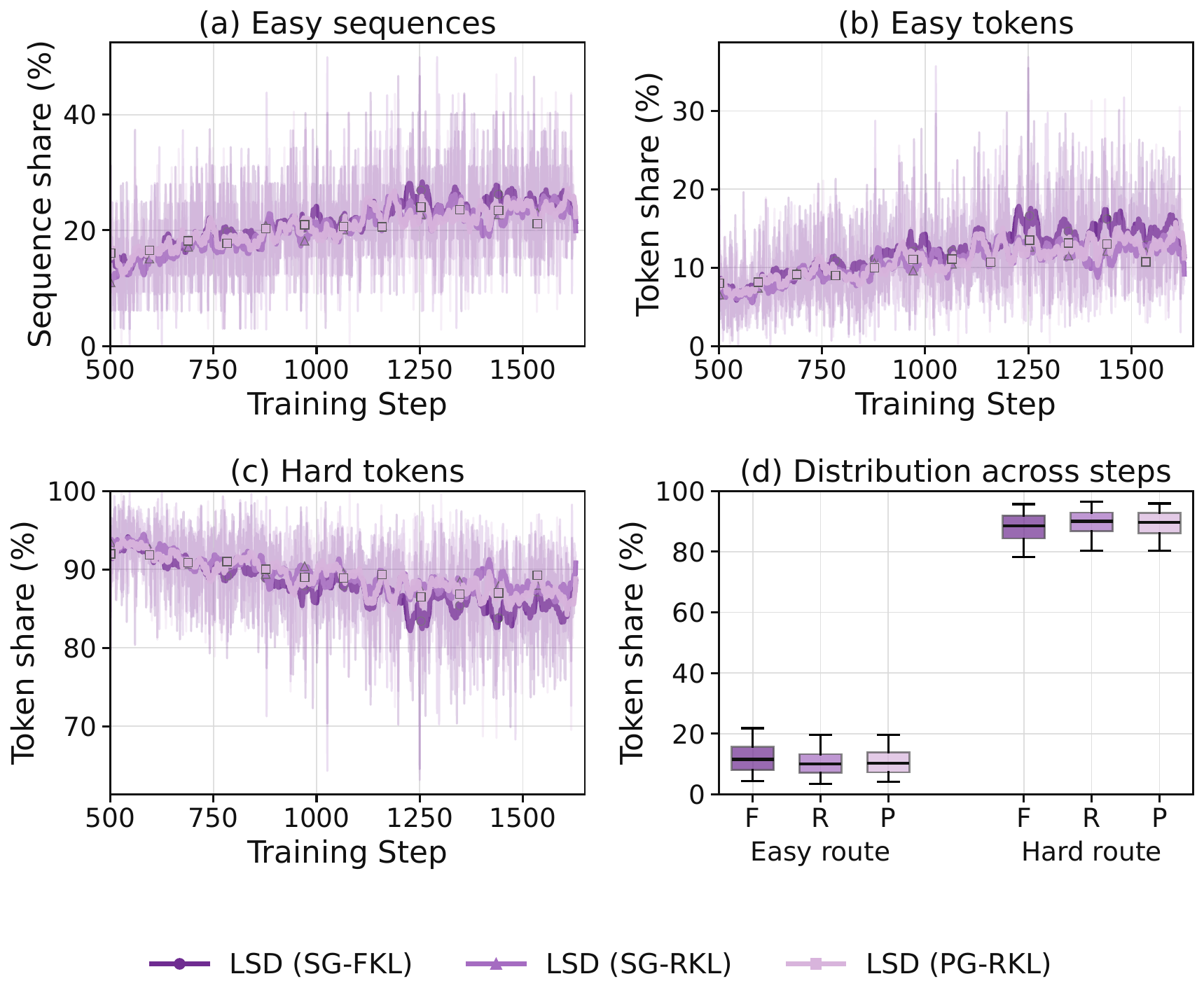}
  \caption{\textbf{Training token allocation between the easy and hard routes.}
  Panels (a)--(c) show raw per-step fractions and centered 21-step means.
  Panel (d) summarizes the distribution of token shares across training
  steps: boxes show the interquartile range and median, and whiskers
  show the 5th and 95th percentiles. F, R, and P denote SG-FKL, SG-RKL,
  and PG-RKL. These are distributions of per-step shares, not
  individual response lengths or uncertainty intervals.}
  \label{fig:training-token-routes}
\end{figure}

The training histories directly record the fraction of sequences and
tokens assigned to each route. Figure~\ref{fig:training-token-routes}
shows their evolution and distribution over the 1128 shared steps
501--1628. The two route fractions sum to one at every recorded step,
up to numerical precision.

\begin{table}[htbp]
  \centering
  \caption{\textbf{Training-route allocation over steps 501--1628.}
  The first three numeric columns are means of logged per-step
  fractions, expressed in percent. The last reports the median and
  interquartile range of the easy-token share.}
  \label{tab:training-token-routes}
  \small
  \resizebox{\linewidth}{!}{%
  \begin{tabular}{lrrrr}
    \toprule
    Variant & Easy seq.\ (\%) & Easy tok.\ (\%) & Hard tok.\ (\%)
      & Easy tok.\ median [IQR] \\
    \midrule
    \methodname{SGFKL}{SG-FKL} & 21.51 & 12.17 & 87.83 & 11.54 [8.09, 15.67] \\
    \methodname{SGRKL}{SG-RKL} & 20.23 & 10.58 & 89.42 &  9.99 [7.11, 13.27] \\
    \methodname{PGRKL}{PG-RKL} & 20.67 & 10.94 & 89.06 & 10.29 [7.31, 13.90] \\
    \bottomrule
  \end{tabular}%
  }
\end{table}

The easy route accounts for a smaller share of tokens than of
sequences, while the hard route retains about 88--89\% of the token
share on average. The values summarize each run's own dynamic routing,
not the fixed easy and hard evaluation cohorts in
Table~\ref{tab:lsd-best-each-max}. The available aggregate logs do not
provide individual response lengths by route, so they do not determine
a per-response length histogram or a token-count-weighted total across
the full training run.

\clearpage
\section{Paired Accuracy and Length on Fixed Easy Sets}
\label{app:paired-preservation}

\paragraph{Evaluation on identical queries.}
We freeze query IDs using RL anchors $b\in\{100,500,1000\}$ and
$\tau=1$: every selected query has 32/32 correct anchor rollouts.

\begin{figure}[!htbp]
  \centering
  \includegraphics[width=0.84\linewidth]{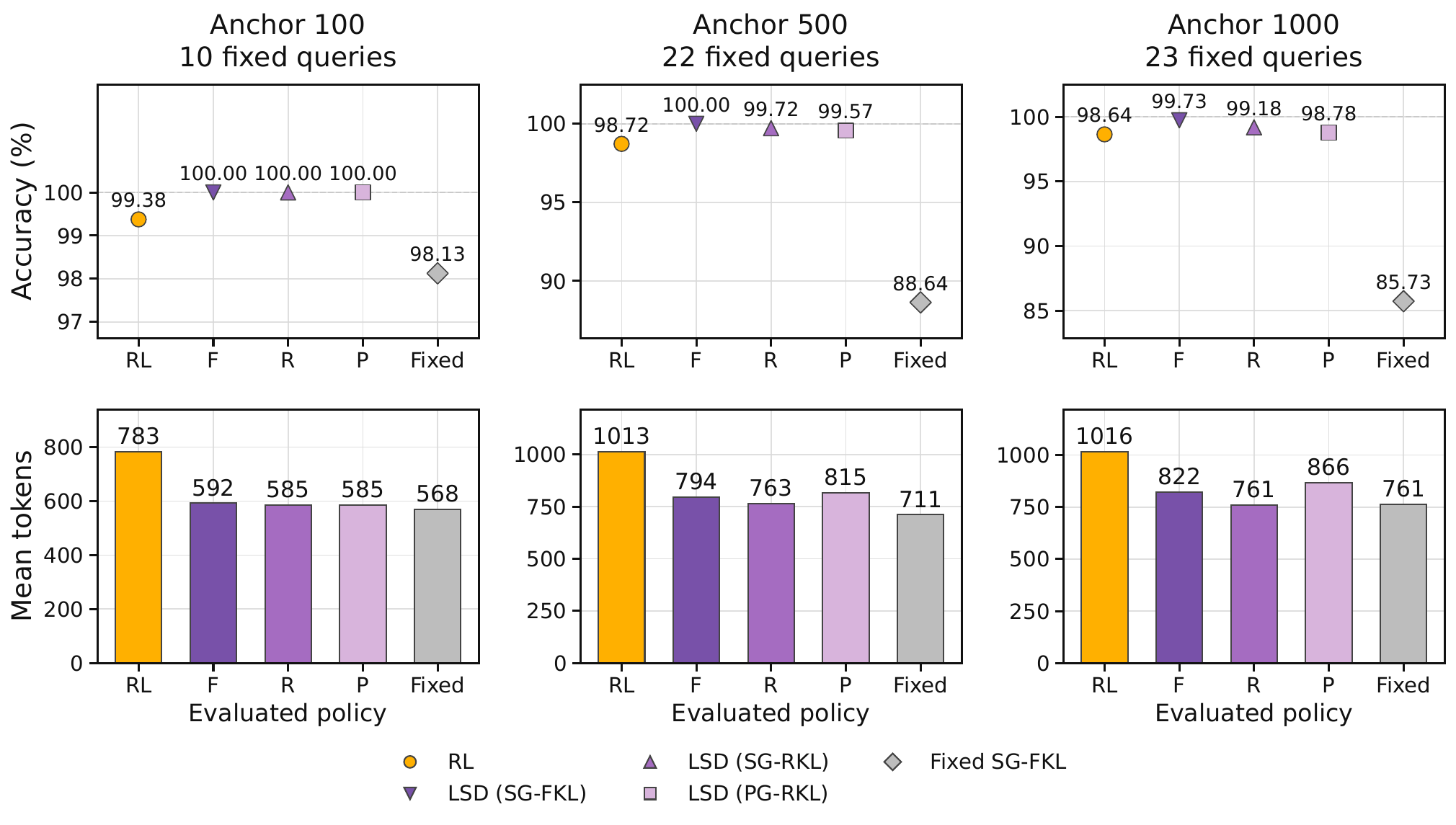}
  \caption{\textbf{Accuracy and length evaluated on the same fixed queries.}
  Each column uses one RL-frozen cohort with $\tau=1$.
  The top row reports empirical accuracy; the bottom row reports mean
  response length. The dashed line marks the 100\% anchor accuracy.
  F, R, and P denote SG-FKL, SG-RKL, and PG-RKL, respectively.
  Every method is evaluated on the
  identical query IDs within a column.}
  \label{fig:paired-preservation}
\end{figure}

\begin{table}[!htbp]
\centering
\small
\caption{\textbf{Paired evaluation on RL-frozen easy sets ($\tau=1$).} $\Delta A$ is the accuracy difference from RL (pp). For $b=1000$, the shared reference is RL step \sharedLSTReferenceStep{}: \sharedLSTReferenceLength{} tokens at 100\% accuracy, the minimum over 345 RL checkpoints.}
\label{tab:paired-preservation}
\resizebox{\linewidth}{!}{%
\begin{tabular}{llrrrr}
\toprule
Anchor & Method & Correct / total & Acc. (\%) & Mean tokens & $\Delta A$ (pp) \\
\midrule
\multirow{5}{*}{$b=100$} & \methodname{RL}{RL} & 318/320 & 99.38 & 783.1 & +0.00 \\
 & \methodname{SGFKL}{LSD (SG-FKL)} & 320/320 & 100.00 & 592.0 & +0.63 \\
 & \methodname{SGRKL}{LSD (SG-RKL)} & 320/320 & 100.00 & 585.3 & +0.63 \\
 & \methodname{PGRKL}{LSD (PG-RKL)} & 320/320 & 100.00 & 585.1 & +0.63 \\
 & \methodname{Fixed}{Fixed SG-FKL} & 314/320 & 98.13 & 568.1 & -1.25 \\
\midrule
\multirow{5}{*}{$b=500$} & \methodname{RL}{RL} & 695/704 & 98.72 & 1013.5 & +0.00 \\
 & \methodname{SGFKL}{LSD (SG-FKL)} & 704/704 & 100.00 & 794.0 & +1.28 \\
 & \methodname{SGRKL}{LSD (SG-RKL)} & 702/704 & 99.72 & 763.0 & +0.99 \\
 & \methodname{PGRKL}{LSD (PG-RKL)} & 701/704 & 99.57 & 815.1 & +0.85 \\
 & \methodname{Fixed}{Fixed SG-FKL} & 624/704 & 88.64 & 710.8 & -10.09 \\
\midrule
\multirow{5}{*}{$b=1000$} & \methodname{RL}{RL} & 726/736 & 98.64 & 1016.2 & +0.00 \\
 & \methodname{SGFKL}{LSD (SG-FKL)} & 734/736 & 99.73 & 822.0 & +1.09 \\
 & \methodname{SGRKL}{LSD (SG-RKL)} & 730/736 & 99.18 & 760.5 & +0.54 \\
 & \methodname{PGRKL}{LSD (PG-RKL)} & 727/736 & 98.78 & 866.1 & +0.14 \\
 & \methodname{Fixed}{Fixed SG-FKL} & 631/736 & 85.73 & 761.4 & -12.91 \\
\bottomrule
\end{tabular}%
}
\end{table}

All three LSD variants use fewer tokens and have higher reported accuracy
than RL on each cohort; all retain 100\% accuracy at $b=100$.

\subsection{The main-table cohort and individual examples}

Table~\ref{tab:paired-main-easy-set} jointly reports accuracy and length
on the easy-query set used in Table~\ref{tab:lsd-best-each-max}, using its
explicitly relaxed RL step-100 cohort ($\tau=0.9$). SG-RKL and PG-RKL
reach 99.43\% and 99.15\% accuracy on these same queries, compared with
98.15\% for RL, while reducing mean length from 997 to 757 and 802
tokens. Fixed produces still shorter responses but lower accuracy,
illustrating why length and correctness must be reported together.
Their hard-query accuracies are 32.50\% and 33.45\%, respectively,
compared with 32.85\% for RL.
SG-FKL reaches 99.10\% and 34.22\% accuracy on the easy and hard
cohorts, with mean lengths of 859 and 2511 tokens, respectively.

Random Routing matches SG-FKL's EMA teacher ($H=4$), objective, coefficient, and per-step distillation group count. Groups are selected uniformly at random, with GRPO retained on all groups.

\begin{table}[htbp]
\centering
\small
\caption{\textbf{Paired accuracy and response length.} Easy and hard columns use the RL step-100 cohort ($\tau=0.9$) and its complement; LSD and Random Routing Hard Acc. is derived from overall and easy-set accuracy. $\mathrm{LST}_{1000}$ uses the strict step-1000 cohort and shared reference in Table~\ref{tab:paired-preservation}.}
\label{tab:paired-main-easy-set}
\setlength{\tabcolsep}{4pt}
\resizebox{\linewidth}{!}{%
\begin{tabular}{lrrrrrr}
\toprule
\multirow{2}{*}{Method} & \multirow{2}{*}{\shortstack{Avg.\\Pass@1 (\%)}} & \multicolumn{2}{c}{Easy queries} & \multicolumn{2}{c}{Hard queries} & \multirow{2}{*}{$\mathrm{LST}_{1000}$ (\%)} \\
\cmidrule(lr){3-4}\cmidrule(lr){5-6}
& & Acc. (\%) & Mean tokens & Acc. (\%) & Mean tokens & \\
\midrule
\methodname{RL}{RL} & 42.90 & 98.15 & 996.9 & 32.85 & 2329.3 & 19.03 \\
\methodname{SGFKL}{LSD (SG-FKL)} & 44.20 & 99.10 & 859 & 34.22 & 2511 & -3.72 \\
\methodname{SGRKL}{LSD (SG-RKL)} & 42.80 & 99.43 & 757.2 & 32.50 & 2395.0 & -10.92 \\
\methodname{PGRKL}{LSD (PG-RKL)} & 43.56 & 99.15 & 801.6 & 33.45 & 2392.5 & 1.45 \\
Random Routing & 22.08 & 89.74 & 895.0 & 9.78 & 2006.6 & -11.00 \\
\methodname{Fixed}{Fixed SG-FKL} & 30.44 & 94.89 & 690.6 & 18.72 & 1322.8 & -10.81 \\
\bottomrule
\end{tabular}%
}
\end{table}

\begin{figure}[htbp]
  \centering
  \includegraphics[width=\linewidth]{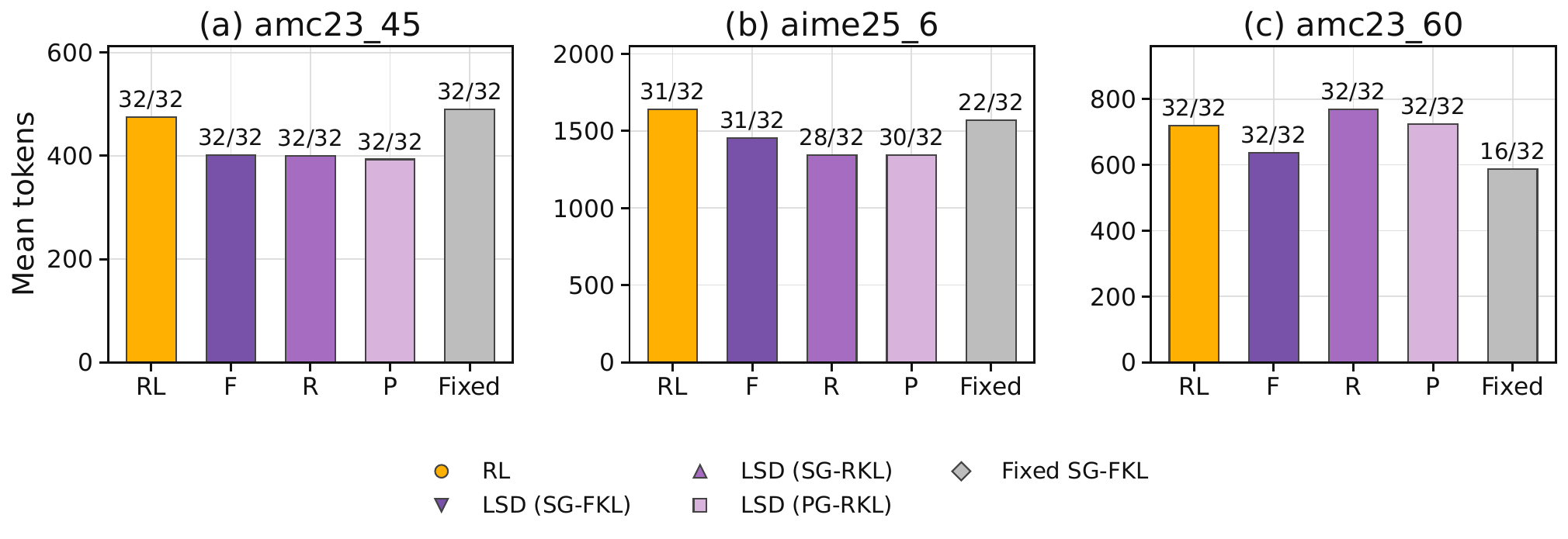}
  \caption{\textbf{Query-level examples from the strict step-1000 cohort.}
  Bar height is mean response length; labels give correct rollouts out
  of 32. All three queries were answered correctly in all 32 RL anchor
  rollouts. The examples include unchanged correctness, an LSD
  regression, and a Fixed regression. F, R, and P denote SG-FKL,
  SG-RKL, and PG-RKL, respectively.}
  \label{fig:paired-query-examples}
\end{figure}

For \texttt{amc23\_45}, RL and all three LSD variants answer 32/32
correctly, while mean length falls from 475.2 to 401, 400.4, and
393.2 tokens under SG-FKL, SG-RKL, and PG-RKL, respectively.
This is a direct example of shortening at identical observed accuracy.
It is selected near the median joint length reduction among queries
with 32/32 correctness for RL, SG-RKL, and PG-RKL and shorter responses
under both RKL variants.

The other examples expose the limits of the aggregate result.
For \texttt{aime25\_6}, correctness falls from RL's 31/32 to 28/32
under SG-RKL and 30/32 under PG-RKL despite shorter responses;
SG-FKL retains 31/32 while reducing mean length from 1640.9 to 1453 tokens.
For \texttt{amc23\_60}, RL and all three LSD variants retain 32/32
correctness; SG-FKL reduces mean length from 720.4 to 637 tokens, while Fixed drops
to 16/32 while shortening its response.
These examples are selected as the largest respective correctness
regressions among shortened queries in the cohort, so that the case
analysis includes failure cases as well as successful preservation.

\clearpage
\section{Related Work}
\label{app:related-work}

Approaches to efficient reasoning broadly include test-time compute allocation, RL-based optimization, and distillation. 

\paragraph{Test-time compute allocation.}
At inference time, reasoning computation can be adjusted through the length of reasoning traces, the number of sampled solutions, and verification or search. Chain-of-thought prompting and self-consistency illustrate the benefits of explicit reasoning and exploring multiple solution paths~\citep{wei2022chain,wang2022self}, while process supervision supports verifier-guided selection~\citep{lightman2024lets}. However, longer reasoning can waste computation on simple problems~\citep{chen2024not}. \citet{snell2024scaling} show that effective compute allocation depends on prompt difficulty, motivating adaptive inference strategies. \citet{wan2026shadow} further formulate cross-query budget allocation using a global shadow price to balance the marginal utility of reasoning computation. Budget forcing provides another way to control the amount of reasoning at test time~\citep{muennighoff2025s1}. These methods adjust how much computation a model spends when answering a query.

\paragraph{RL-based reasoning efficiency.}
RL can train policies to use computation more efficiently. L1 optimizes accuracy together with adherence to requested reasoning-length constraints~\citep{aggarwal2025l1}, and lazy length penalties incorporate response-length reduction into reasoning RL~\citep{yuan2026shorten}. DAST uses difficulty-dependent budgets, reward shaping, and preference optimization to discourage excessive reasoning on easier problems while retaining sufficient computation for harder ones~\citep{shen2025dast}. AdapThink adapts length penalties to query difficulty~\citep{xu2026adapthink}. Training efficiency can also be improved through sample selection: DAPO filters rollout groups with uninformative rewards~\citep{yu2025dapo}, while online difficulty filtering focuses learning on tasks of intermediate difficulty~\citep{bae2026online}. Under group-relative objectives, however, all-correct groups have no reward-based relative advantage~\citep{shao2024deepseekmath}. This leaves little direct signal for preserving their existing concise behavior as the policy continues learning from other prompts.

\paragraph{Distillation and LSD.}
Distillation provides token-level supervision for learning concise reasoning. Generalized Knowledge Distillation trains students on their own generated sequences using teacher feedback, supports different divergence objectives, and can be combined with RL fine-tuning~\citep{agarwal2024onpolicy}. CRISP uses a periodically refreshed student copy conditioned on a conciseness instruction as its teacher and optimizes reverse KL on all student rollouts~\citep{sang2026opsdc}. Contrastive On-Policy Distillation compares teacher probabilities under light- and heavy-reasoning instructions to construct token-level advantages~\citep{ruan2026contrastive}.

Unlike these approaches, LSD combines on-policy distillation with online difficulty routing. Its focus is preserving concise behavior on already-solved queries during ongoing RL training, which we evaluate through LST on fixed easy-query sets.

\clearpage
\section{Experimental Configurations}
\label{app:experimental-configurations}

Table~\ref{tab:task-configurations} consolidates shared settings and task-specific
differences. Table~\ref{tab:method-configurations} specifies the RL and LSD variants.

\begin{table}[!htbp]
  \centering
  \caption{\textbf{Training and inference configurations for RL and LSD.} Values spanning both
  columns are shared. Agent mini-batches count transformed training rows.
  A dash indicates an unspecified setting.}
  \label{tab:task-configurations}
  \small
  \setlength{\tabcolsep}{4pt}
  \begin{tabularx}{\linewidth}{@{}lXX@{}}
    \toprule
    Parameter & Single-turn reasoning & Multi-turn agentic tasks \\
    \midrule
    Optimizer & \multicolumn{2}{c}{AdamW; $\beta=(0.9,0.999)$} \\
    Learning rate / weight decay & \multicolumn{2}{c}{$10^{-6}$ / $0.01$} \\
    Training batch & \multicolumn{2}{c}{32 prompts $\times$ 8 rollouts = 256 trajectories} \\
    PPO epochs / advantage & \multicolumn{2}{c}{1 / GRPO} \\
    PPO clip $(\epsilon_{\mathrm{low}},\epsilon_{\mathrm{high}})$ & \multicolumn{2}{c}{$(0.20,0.28)$} \\
    Entropy / RL KL coef. & \multicolumn{2}{c}{0 / 0} \\
    Sampling / top-$k$ / TP & \multicolumn{2}{c}{Enabled / $-1$ / 1} \\
    \midrule
    Backbone & Qwen3-4B-Base & Qwen3-8B-Base \\
    Training data & DAPO-Math-17K & CutTheBill \texttt{train} \\
    Evaluation data & AMC23 / AIME25 / AIME26 (83 / 30 / 30 queries) & BrowseComp-Plus \texttt{test} (830 queries) \\
    Agent / tools & Single-turn math agent / none & DeepResearch / retrieval + Refine \\
    Max.\ prompt / response tokens & 4096 / 4096 & 700 / 20,000 \\
    Max.\ interaction turns & 1 & 48 \\
    Stopping & EOS or response-token limit & EOS, response-token limit, or turn limit \\
    PPO mini-batch & 64 trajectories & 2048 transformed rows \\
    Dynamic micro-batch limit & 16,384 tokens/GPU & 24,576 tokens/GPU \\
    Dual-clip $c$ & 3 & 10 \\
    LR schedule / warmup & Constant / 0 & Constant / 10 steps \\
    Loss aggregation & \texttt{seq-mean-token-mean} & \texttt{token-mean} \\
    Train $(T,\mathrm{top}\text{-}p,n)$ & $(0.6,1,8)$ & $(1.0,1,8)$ \\
    Validation $(T,\mathrm{top}\text{-}p,n)$ & $(0.6,1,32)$ & $(1.0,0.7,8)$ \\
    Inference engine & vLLM & SGLang \\
    Max.\ model length & 8192 & 20,700 \\
    GPU memory utilization & 0.8 & 0.65 \\
    Runtime options / max.\ sequences & -- / -- & Eager; overlap scheduling off / 128 \\
    Validation / saving interval & -- & Every 25 steps \\
    Parallelism / model precision & FSDP / BF16 & FSDP2 / FP16 \\
    EMA accumulation precision & FP32 (EMA teachers only) & -- \\
    Hardware & 1 node $\times$ 8 GPUs & 1 node $\times$ 8 H20 GPUs \\
    \bottomrule
  \end{tabularx}
\end{table}

\begin{table}[!htbp]
  \centering
  \caption{\textbf{RL and LSD configurations.} LSD retains GRPO on the hard
  route; the gradient row describes only the distillation route. The three
  online LSD variants share these settings across tasks. Fixed SG-FKL
  results are reported for single-turn reasoning.}
  \label{tab:method-configurations}
  \small
  \setlength{\tabcolsep}{3pt}
  \begin{tabular}{@{}lccccc@{}}
    \toprule
    Parameter & \methodname{RL}{RL} & \methodname{Fixed}{\shortstack{Fixed\\SG-FKL}} & \methodname{SGFKL}{SG-FKL} & \methodname{SGRKL}{SG-RKL} & \methodname{PGRKL}{PG-RKL} \\
    \midrule
    Teacher & -- & Frozen $\pi_0$ & \multicolumn{3}{c}{EMA} \\
    Routing & -- & Fixed map & \multicolumn{3}{c}{Online rollout-group accuracy} \\
    Routing threshold $\tau$ & -- & \multicolumn{4}{c}{1.0} \\
    EMA half-life $H$ & -- & -- & \multicolumn{3}{c}{4} \\
    Distillation objective & -- & \multicolumn{2}{c}{Top-32 FKL} & Top-32 RKL & Sample-token RKL \\
    Distillation gradient & -- & \multicolumn{3}{c}{Supervised} & Policy gradient \\
    LSD coefficient & -- & \multicolumn{4}{c}{1.0} \\
    Easy-route RL coefficient & -- & \multicolumn{4}{c}{0} \\
    PG advantage clip & -- & -- & -- & -- & 10 \\
    \midrule
    Single-turn W\&B ID & \texttt{s71jbnha} & \texttt{tuabqb6l} & \texttt{z7hhhn5x} & \texttt{kvjksgpr} & \texttt{178n91ko} \\
    \bottomrule
  \end{tabular}
\end{table}

\end{document}